\documentclass[runningheads]{llncs}
\usepackage{graphicx}
\usepackage{amsmath,amssymb} %

\usepackage[utf8]{inputenc} %
\usepackage[T1]{fontenc}    %
\usepackage{hyperref}   %
\usepackage{url}            %
\usepackage{booktabs}       %
\usepackage{amsfonts}       %
\usepackage{nicefrac}       %
\usepackage{microtype}      %
\usepackage{xcolor}         %
\usepackage[numbers,sort]{natbib} %
\usepackage{floatrow}
\usepackage{subfig}
\usepackage{newtxmath}
\usepackage{caption}
\usepackage{graphicx}
\usepackage{pifont}%
\usepackage{fontawesome}
\usepackage{multirow}
\usepackage{xspace}
\usepackage{overpic}
\usepackage{comment}
\usepackage{amsmath}

\newcommand{\pp}[1]{\newline\noindent\textbf{#1}}

\newcommand{\xmark}{\ding{55}}%
\newcommand{\bs}[1]{\boldsymbol{#1}}
\newcommand{\compressor}{{\em neural compressor}\xspace}
\newcommand{\augmentor}{{\em augmentation network}\xspace}
\newcommand{\bfcompressor}{{\bf Neural compressor}\xspace}
\newcommand{\bfaugmentor}{{\bf Augmentation network}\xspace}

\hypersetup{
     colorlinks   = true,
     citecolor    = blue
}

\usepackage{amsmath,amsfonts,bm}

\newcommand{\test}{\mathcal{D_{\mathrm{test}}}}

\def\eqref#1{equation~\ref{#1}}

\def\1{\bm{1}}

\DeclareMathAlphabet{\mathsfit}{\encodingdefault}{\sfdefault}{m}{sl}
\SetMathAlphabet{\mathsfit}{bold}{\encodingdefault}{\sfdefault}{bx}{n}

\newcommand{\etens}[1]{\mathsfit{#1}}

\def\etR{{\etens{R}}}

\def\etX{{\etens{X}}}

\begin{document}
\mainmatter

\title{Compressed Vision for \\ Efficient Video Understanding}
\titlerunning{Compressed Vision}
\authorrunning{O. Wiles et al.}

\author{
    Olivia Wiles*\and 
    Jo\~ao Carreira \and
    Iain Barr \and
    Andrew Zisserman \and \\
    Mateusz Malinowski*
}

\institute{DeepMind, London, U.K.\\
* \email{\{oawiles,mateuszm\}@deepmind.com}
}

\maketitle

\begin{abstract}
Experience and reasoning occur across multiple temporal scales: milliseconds, seconds, hours or days. 
The vast majority of computer vision research, however, still focuses on individual images or short videos lasting only a few seconds.
This is because handling longer videos require more scalable approaches even to process them.
In this work, we propose a framework enabling
research on hour-long videos with the same hardware that can now process second-long videos. 
We replace standard video compression, e.g. JPEG, with neural compression and show that we can directly feed  compressed videos as inputs to regular video networks.
Operating on compressed videos improves efficiency at all pipeline levels -- data transfer, speed and memory -- making it possible to train models faster and on much longer videos. Processing compressed signals has, however, the downside of precluding standard augmentation techniques if done naively. We address that by introducing a small network that can apply transformations to latent codes corresponding to commonly used augmentations in the original video space. 
We demonstrate that with our compressed vision pipeline, we can train video models more efficiently on popular benchmarks such as Kinetics600 and COIN.
We also perform proof-of-concept experiments with new tasks defined over hour-long videos at standard frame rates. Processing such long videos is impossible without using compressed representation. 
\end{abstract}

\section{Introduction}

\begin{figure}[h]
    \centering
    \subfloat[]{\includegraphics[width=0.4\linewidth]{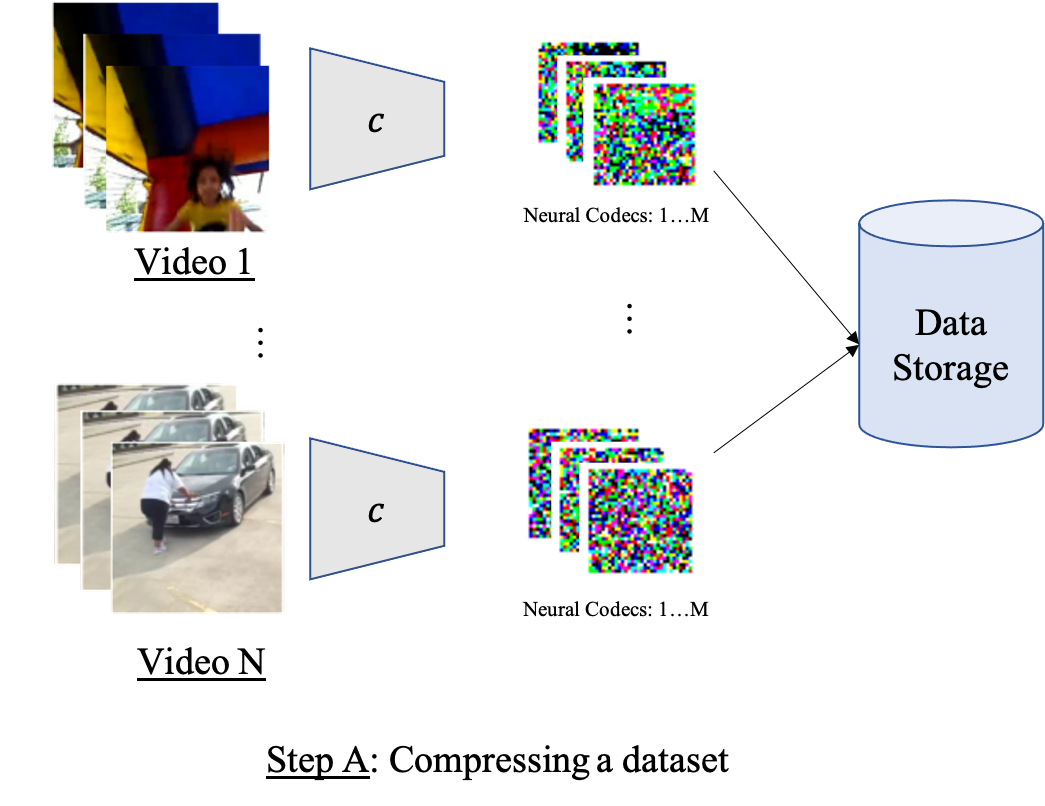}\label{fig:compression}}
    \hspace{1em}
    \subfloat[]{\includegraphics[width=0.55\linewidth]{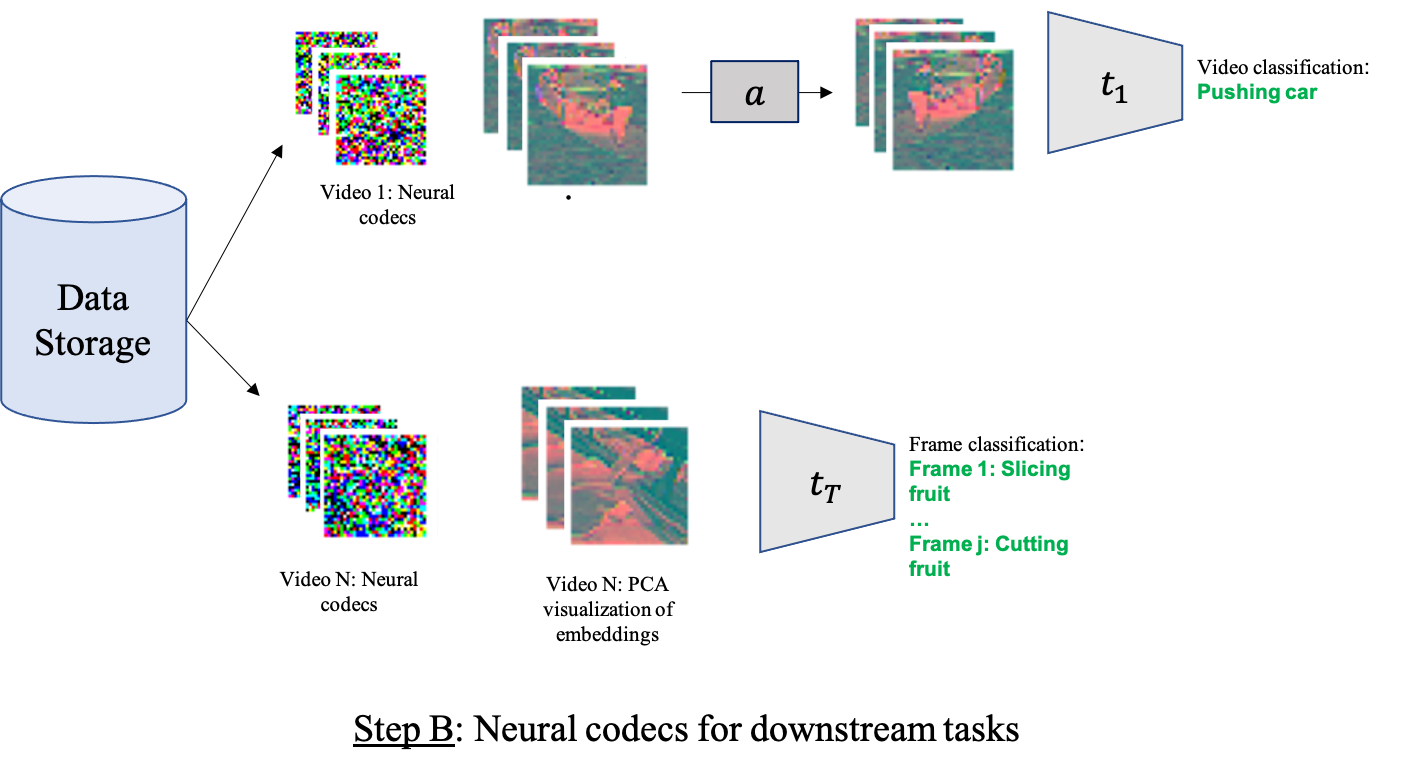}\label{fig:downstream}}
    \caption{{\bf The compressed vision pipeline.} Videos are first compressed using a \compressor $c$ to produce codes. These are stored on a disk and the original videos can be discarded. The neural codes are directly used to train video tasks $t_1 \dots t_T$. We can optionally augment these codes with augmented versions using an \augmentor $a$ (here we show a flipping augmentation). Note that within our framework, all the computations are done in a more efficient compressed space as decompression is not required at any stage of the pipeline.
    }
    \label{fig:overviewfigure}
\end{figure}

Most computer vision research focuses on short time scales of two to ten seconds at 25fps (frames-per-second) because vision pipelines do not scale well beyond that point.
Raw videos are enormous and must be stored compressed on a disk; after loading them from a disk, they are decompressed and placed in a device memory before using them as inputs to neural networks.
In this setting, and with current hardware, training models on minute-long raw videos can take prohibitively long or take too much physical memory. Even loading such videos onto GPU or TPU might become infeasible, as it requires decompressing and transferring, often over the bandwidth-limited network infrastructure. While previous work has considered using classic video or image compressed codes (such as JPEG or MPEG) directly as input to their models~\cite{gueguen2018faster,ehrlich2019deep,wu2018compressed,xu2020learning}, this generally requires specialised neural network architectures.

In this work, we propose and investigate a new, efficient and more scalable video pipeline -- {\em compressed vision} -- which preserves the ability to use most of the state-of-the-art data processing and machine learning techniques developed for videos.
The pipeline, described in more detail in Section~\ref{sec:compressed_vision}, has three components. First, we train a \compressor to compress videos. Second, we optionally use \augmentor to transform the compressed space for doing augmentations. Third, we {\em directly} apply standard video backbone architectures
on these neural codes to train and evaluate on standard video understanding tasks (thereby avoiding costly decompression of the videos). As our framework is modular, each component could be replaced with a more efficient variant.

Performing augmentations (e.g.~{\em spatial cropping} or {\em flipping}) is an important component of many pipelines used to train video models but are impossible to perform directly in the compressed space.
Therefore, we face the following dilemma. We either give up on augmentations or we decompress the codes and do the transformations in the pixel space. However, if we decide on the latter, we loose some benefits of the compressed space. Decompressed signals expands the space and if they are long enough they cannot fit to a GPU or TPU memory anymore. Moreover, even though a \compressor yields superior quality at higher compression rates to JPEG or MPEG, it has large decoders that take even more time and space; i.e., neural decompression is  slow.

To overcome the last challenge, we propose an \augmentor\;-- a small neural network that acts directly on latent codes by transforming them according to some operation. 
In brief, the \augmentor for spatial cropping takes crop coordinates and a tensor of latent codes as inputs. Next, it outputs the modified latents that approximate the ones obtained by spatially cropping the video frames.
Note that, unlike \cite{patrick2021space},  we {\em learn} how to augment in the compressed space as opposed to cropping the compressed tensor. 
As a result, we can train an \augmentor for a wider variety of augmentations, such as changing the {\em brightness} or {\em saturation} or even performing {\em rotations}; all these are difficult or impossible by directly manipulating the tensor.

Our approach has the following benefits.
First, it allows for standard video architectures to be directly applied on these neural codes, as opposed to devising new architectures, as in~\cite{wu2018compressed} which trains networks directly on MPEG representations. Second, as demonstrated in Section~\ref{sec:compressed_representation}, we can apply augmentations directly on the latent codes {\em without} the need to decompressing them. This significantly impacts training time and saves memory.
With these two properties, we can use standard video pipelines with minimal modifications and achieve competitive performance to operating on raw videos (RGB values).

In summary, we show that neural codes generalise to a variety of datasets and tasks (whole-video and frame-wise classification), and are better at compression than JPEG or MPEG.
To enable augmentations in the latent space, we train and evaluate a separate network, which conditioned on transformation arguments, outputs transformed latents.
We also demonstrate our framework on much longer videos.
Here, we collected a large set of ego-centric videos recorded by tourists walking in different cities\footnote{The reader can get a feel for the dataset here \url{https://youtu.be/MIzp8Wrj44s?t=131}.}. These videos are long and continuous, and last between 30 minutes to ten hours. One can see a few samples of our results on the  website\footnote{Project website: \url{https://sites.google.com/view/compressed-vision}}. We plan to update it in the future, e.g., to include the source code.

\section{Related Work}

Our work is built upon the following prior work.
\pp{Operating on a compressed space.}
Directly using compressed representations for downstream tasks for video or image data has primarily been studied by considering standard image and video codecs such as JPEG or MPEG~\cite{gueguen2018faster,ehrlich2019deep,wu2018compressed}, DCT~\cite{xu2020learning,nash2022transframer} or scattering transforms~\cite{oyallon2018compressing}. 
However, in general these approaches require devising novel architectures, data pipelines, or training strategies in order to handle these representations.
Another strategy is to learn representations that are {\em invariant} to a range of transformations, as in \cite{dubois2021lossy}.
However, this requires a-priori knowledge of the downstream tasks to be invariant to.
In our case, we do not have this knowledge, as we want to have the same representation to be used for a variety of potentially unknown tasks.
\pp{Discrete representations for compression.} Some modalities such as language are inherently discrete and naturally benefit from neural discrete representations~\cite{mnih2014neural,oord2017neural}.
However, the same representation has become increasingly more common in generative modelling of images~\cite{esser2020taming,ramesh2021zero} and videos~\cite{walker2021predicting,yan2021videogpt}. 
In our work, we demonstrate a different use of vector quantization: to compress videos. Thus, we can directly train classifiers on the resulting compressed space, without the need for decompression, making one-hour long training and inference feasible. 
\pp{Augmentations using latents.} 
In the image domain, it has been demonstrated that augmentations in an embedded space learned via a GAN~\cite{chai2021ensembling,jahanian2020steerability,harkonen2020ganspace} or an encoder-decoder model~\cite{devries2017dataset} can be used to improve classification performance.
In the video setting,~\cite{patrick2021space} have applied augmentations directly in a latent space, but only considered cropping of the tensor -- an operation that can readily be designed. Instead, we propose a learnable approach to approximate augmentations in a latent space, so that we can generalize the approach to arbitrary augmentations.
\pp{Video understanding.} 
Understanding long videos encompasses many important sub-problems including action or event understanding and reasoning over longer spatio-temporal blocks. It is also computationally demanding.
On these tasks, transformer-like architectures~\cite{bello2019attention,bertasius2021space,fan2021multiscale,huang2019ccnet,vaswani2017attention,wang2018non} could potentially compete with spatio-temporal convolutional networks (3D CNNs)~\cite{carreira2017quo,feichtenhofer2019slowfast,stroud2020d3d,tran2015learning,xie2017rethinking} that reason about time and space locally. However, videos in current datasets are often too short~\cite{kay2017kinetics,gu2018ava,kuehne2011hmdb,sigurdsson2018charades,soomro2012ucf101} and thus these data hungry architectures are not so useful. This work demonstrates how neural compression can be used to scale architectures to operate on much longer sequences.
\pp{Long sequence processing.}
Long-Range Arena~\cite{tay2020long} was designed to challenge the ability of transformer-like architectures to model long-term dependencies and their efficiency. 
These problems are well handled directly by efficient transformers~\cite{kitaev2020reformer,ramesh2021zero,wang2020linformer,zaheer2020big}. Walking Tours also sets up a benchmark for efficient and long-term video understanding by probing networks on their ability to handle long sequences where input signals are high-dimensional.

\section{Compressed Vision}
\label{sec:compressed_vision}
Here, we introduce  our {\em compressed vision} pipeline. Next, we discuss each component of the pipeline.

\begin{figure}[h]
    \centering
    \includegraphics[width=0.9\linewidth]{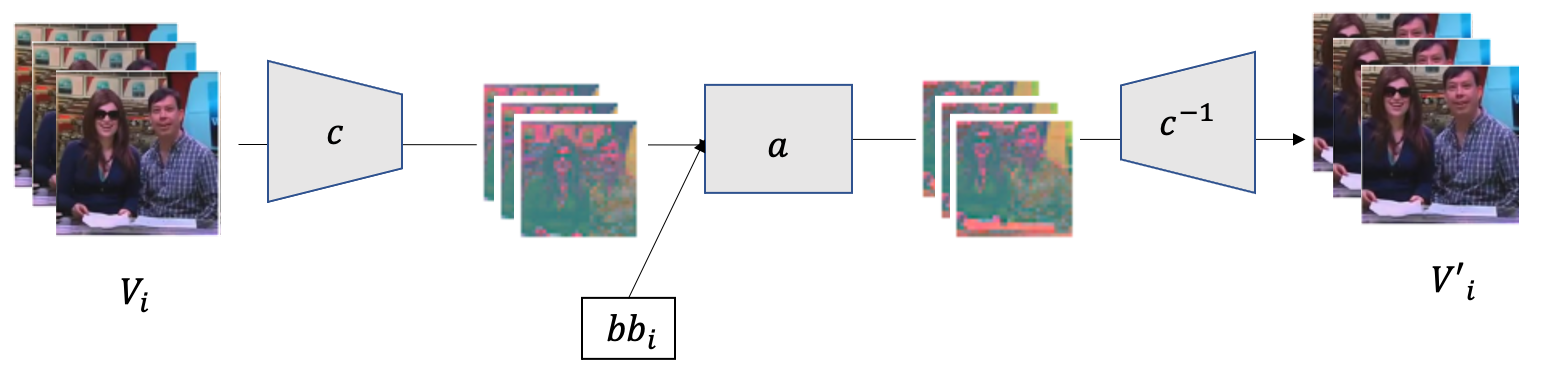}
    \caption{{\bf Augmentation Network.} We train a network $a$ that, conditioned on the bounding box coordinates of the desired spatial crops, performs that spatial crop directly on the latent codes (these embeddings are visualised using PCA). As shown here, after decompression, the resulted video corresponds to spatially-cropped original video.
    Even though cropping is our running example, our methodology extends to potentially arbitrary augmentations.
    }
    \label{fig:equivariant}
\end{figure}

\subsection{Our Pipeline}
The typical setup used for training video networks is motivated by that videos are stored in a compressed form.
To train neural networks, these videos need to be loaded into a memory and decompressed; taking up space and time.

Instead, it would be preferable to operate directly on the neural codes.
Other methods that have investigated this setup used the MPEG encoding directly~\cite{wu2018compressed}. However, this requires devising specific architectures and pipelines to handle compressions in that form.
As a result, a significant amount of work and progress in the last ten years to apply neural networks to video data cannot be directly applied to these representations.
Instead, our pipeline creates compressed tensors that are amenable to being loaded efficiently and passed directly as input to standard video pipelines with minimal, if any, changes in hyper-parameters.

An overview of our pipeline is given in Figure~\ref{fig:overviewfigure}.
It operates in three stages.
First, we have a \compressor $c$ that is used to compress videos a-priori and store them on a disk or other data carriers (Figure~\ref{fig:compression}).
After learning this network, there is no further need to touch the original videos and in fact they can be removed to free-up space.
Second, for any downstream task, we have a downstream network $t_i$ that is applied to these neural codes to solve that task (Figure~\ref{fig:downstream}).
These tasks may be varied (e.g.~frame prediction, reconstruction, classification, etc.) but they all use the same neural codec.
Additionally, as we use spatial tensors, many standard architectures can be used while only changing hyper-parameters of the model.
Finally, we also learn an \augmentor $a$ that (Figure~\ref{fig:equivariant}), conditioned on a neural code $\etR$, can transform it into a new representation $\etR' = c(\etX)$. 
When decompressed using a decoder $c^{-1}$, $\etR'$ approximates an augmentation transformation $A$, e.g. a spatial crop, acting on the corresponding rgb-valued video $\etX$. That is, we are aiming at
$c^{-1} (a (c(\etX))) \approx A (\etX)$.

\subsection{Learning Neural Codecs}
\label{sec:compressed_representation}

To learn $c$, we build on a standard VQ-VAE encoder-decoder model~\cite{oord2017neural}.
The VQ-VAE model uses an encoder to map images into a spatial tensor. Next, each vector in the spatial tensor
is compared with a sequence of embeddings using nearest neighbours. Indices that correspond to the chosen embeddings are stored as codes.
These codes represent the input images, which can now be discarded: we need only store codes for each image and the fixed-length sequence of embeddings, which we call the codebook.
To decode, the codes are mapped to the corresponding embeddings from the codebook, and passed to the decoder. 
To extend such a representation to videos simply requires modifying the encoding and decoding networks.
Instead of using 2D spatial convolutions, we use 3D convolutions to obtain a 3D tensor of codes.
We can additionally use additional codebooks for improved performance at the cost of using more memory.

\paragraph{Discussion.} There are three factors controlling compression rate.
First, the size $T_T \times T_H \times T_W$ of the tensor learned by the encoder controls its spatio-temporal size. 
Second, the number of codebooks $T_C$ controls the number of codes stored at each location in the tensor.
Finally, the number of codes $K$ in the codebook controls the number bits required to store a code.
An RGB-video, $I_T\times I_H \times I_W \times 3$, gives a compression rate:
$c_r = \frac{I_T I_H I_W * 3 * \log_2 256}{T_T T_H  T_W T_C \log_2 K} $. Although we focus on compressing space, we provide promising time-compression results in the supplementary. 

\paragraph{Training general representations.}
We train the discrete codes using an auto-encoding task and $l_1$ reconstruction loss.
The codes should encode a neural codec that approximates the original video while removing redundancy.
As a result, we will be able to use it for a variety of (potentially unknown) downstream tasks (e.g.~classification).
Moreover this representation should be useful for finding augmentations in that space as different videos (including augmented ones) will map to a separate representation.
Note that the intuition here is {\em not} that we are learning an invariant representation as in \cite{dubois2021lossy}, which would impede the ability to learn augmentations in the representation space.

\paragraph{Encoder/decoder architecture.}
First, we create spatio-temporal patches of the videos. 
Next, we use strided 3D CNNs with inverted ResNet blocks~\cite{sandler2018mobilenetv2} to obtain a spatial tensor of vectors.
We independently quantize the resulting vectors.
This gives intermediate representations that are reversed using strided transposed convolutions for decoding. 
Further details about the architecture and hyperparameters are given in the supplementary material. 

\subsection{Video Tasks with Neural Codes}
\label{sec:compressed_downstream}
Our final aim is to use the neural codes as the input for arbitrary video tasks.
As our neural codes have a spatio-temporal structure, we can train standard modern architectures, such as S3D \cite{xie2017rethinking}, directly on this representation for the downstream task.
During training, we sample codes that correspond to videos, and obtain the corresponding embeddings using the codes as indices.
We input these tensors to our downstream model, e.g. S3D~\cite{xie2017rethinking}. To maintain the same spatial resolution as the original S3D, we modified the strides of the convolutions.
For example, for a compressed tensor of size $28\times28$ we use a stride of $1$ (versus $2$ for a $224\times224$ image).

\subsection{Augmentations in the Compressed Space}
\label{sec:equivariant_representation}

In a standard vision pipeline, being able to augment the input at train and evaluation time often leads to large boosts in performance.
Here we describe how a similar procedure can be applied to our neural codes for similar gains.

The central idea is to learn to approximate augmentations but in the compressed space. We use a relatively small neural network, which we call an \augmentor, to learn such an approximation.
Let $A$ be an augmentation transformation on the input video $\etX$. For instance, $A(\etX)$ can spatially crop the input signal $\etX$. Given $\etX$ and the bounding box $\textit{bb}$
describing the coordinates of the crop, we train a neural network $a(\cdot)$ to perform the equivalent transformation but in the compressed space. If $c$ and $c^{-1}$ denote encoder and decoder, we want to maintain the following relationship: $A_{\textit{bb}}(\etX) = c^{-1}(a(c(\etX), \textit{bb}))$.

\paragraph{Downstream tasks.}
As in a traditional pipeline, we can apply augmentations at both train and test time to boost performance.
In our pipeline, we proceed as follows and use an \augmentor trained, for example, to predict spatial crops.
At train time, for each compressed video clip $\etX$ in a batch, we randomly select a bounding box $\textit{bb}$ and then apply the transformation $a(\etX, \textit{bb})$ to obtain a transformed version of the video clip $\etX'$ which is the input to the downstream network.
At test time, for a given compressed video clip $\etX$, we linearly interpolate $N$ bounding box coordinates.
Each of these are used to transform the video clip to create $N$ augmented versions of each video clip.
The predictions for all $N$ clips are averaged to obtain the final prediction for that clip.

\paragraph{Training.}
We train the \augmentor after learning the neural codec (so there is no requirement that these steps are done on the same datasets) and keep $c$ and $c^{-1}$ fixed.
To train, we select a given augmentation class (such as spatial cropping) which is parameterised by some set of values (such as the bounding box for spatial cropping).
We then create training pairs by randomly selecting pairs: a video $\etX$ and a bounding box $\textit{bb}$.
We apply the augmentation to the video to get its augmented version $\etX'$.
Finally, we minimize the following reconstruction loss (we use an $l_1$ loss) to train the corresponding transformation network:
\begin{equation}
|| A(\etX) - c^{-1}(a(c(\etX), \textit{bb})) ||_{l_1}.
\end{equation}
We find that using an $l_1$ loss is sufficient for good results on downstream tasks; we do not require more complicated training pipelines that use adversarial losses.
Note that we only train $a$ and all other transformations or networks are frozen.
$a(\cdot)$ can then be applied to any video to simulate the given augmentation.

\paragraph{Implementation.}
We use a multi-layer perceptron (MLP) with three hidden layers and a two-layer transformer \cite{dosovitskiy2020image} to represent $a(\cdot)$. 
The MLP is applied to a tensor representing an external transformation,  e.g.,  bounding box coordinates or binary values describing when flipping. This creates a representation that we condition on.
Obtained features have the same number of channels as the neural codes.
We then broadcast them over the spatial and temporal dimensions of the compressed tensor and concatenate along the channel dimension.
The result is passed through the transformer to obtain a transformed representation of the same dimension as the neural codes but conditioned on the external transformation.
Please see the supplementary material for the precise details.

\begin{figure}[h]
    \centering
    \tiny
    \subfloat{
    \begin{tabular}[b]{c}
    \begin{overpic}[width=0.32\linewidth,trim={.5cm 3cm 7.6cm 3cm},clip]{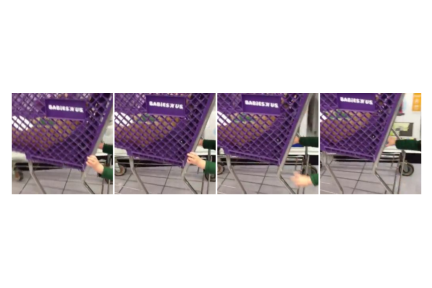}
    \put(-3,30){\makebox(0,0){\rotatebox{90}{Video Frames}}}
    \end{overpic}
    \includegraphics[width=0.32\linewidth,trim={.5cm 3cm 7.6cm 3cm},clip]{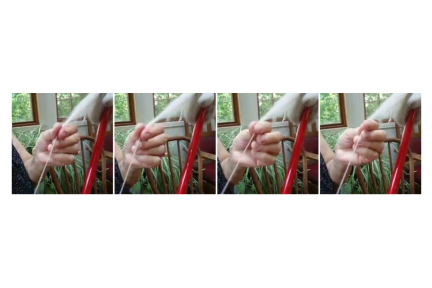} 
    \includegraphics[width=0.32\linewidth,trim={.5cm 3cm 7.6cm 3cm},clip]{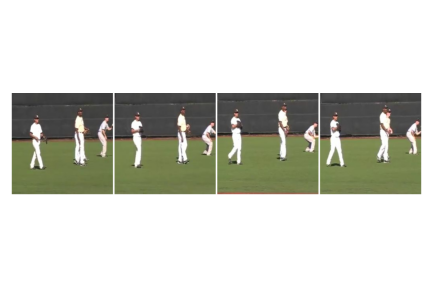}
    \end{tabular}}
    
    \vspace{-3em}
    
    \subfloat{
    \begin{tabular}[b]{c}
    \begin{overpic}[width=0.32\linewidth,trim={.5cm 3cm 7.6cm 3cm},clip]{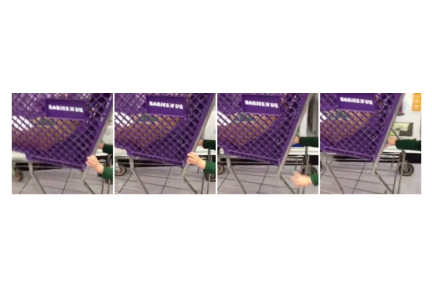}
    \put(-3,30){\makebox(0,0){\rotatebox{90}{CR: 30x}}}
    \end{overpic}
    \includegraphics[width=0.32\linewidth,trim={.5cm 3cm 7.6cm 3cm},clip]{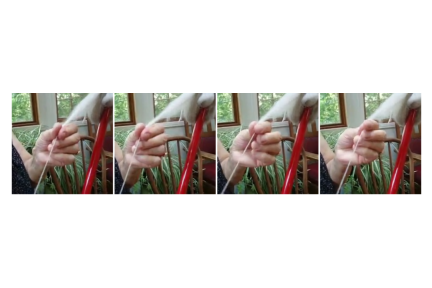}
    \includegraphics[width=0.32\linewidth,trim={.5cm 3cm 7.6cm 3cm},clip]{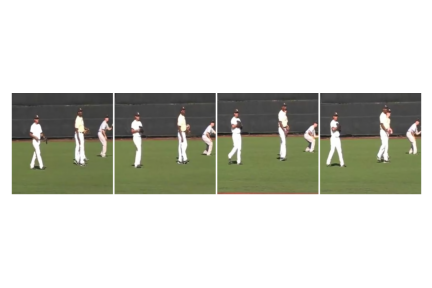}
    \end{tabular}}
    
    \vspace{-3em}
    
    \subfloat{
    \begin{tabular}[b]{c}
    \begin{overpic}[width=0.32\linewidth,trim={.5cm 3cm 7.6cm 3cm},clip]{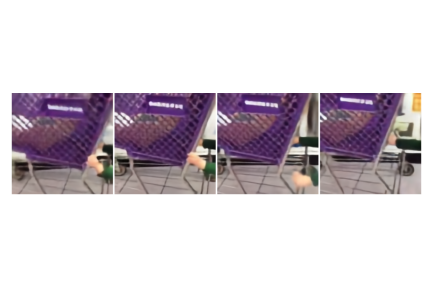}
    \put(-3,30){\makebox(0,0){\rotatebox{90}{CR: 475x}}}
    \end{overpic}
    \includegraphics[width=0.32\linewidth,trim={.5cm 3cm 7.6cm 3cm},clip]{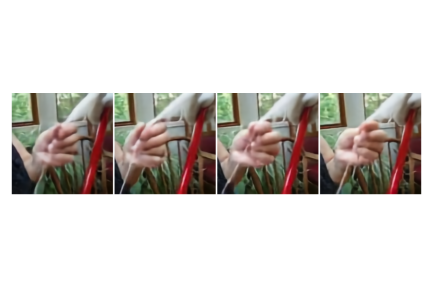}
    \includegraphics[width=0.32\linewidth,trim={.5cm 3cm 7.6cm 3cm},clip]{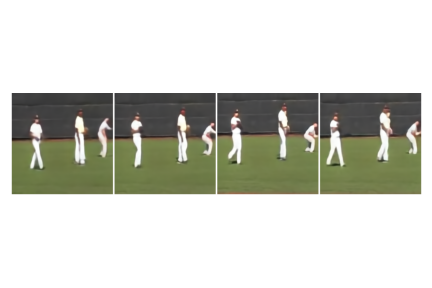}
    \end{tabular}}
    \caption{{\bf Reconstruction results at different compression levels.} At 30x compression (second row), the results are virtually indistinguishable from the original videos (first row). At 475x compression (third row), the codes lose higher frequency detail but still capture the overall structure of the scene.}
    \label{fig:reconstruction}
\end{figure}

\section{Experiments}
\label{sec:experiments}
Here, we evaluate our generic pipeline using compressed video representations with latent augmentations.
We  test three things: (1) the utility of our neural codes for achieving high quality performance on a variety of downstream tasks; (2) the memory and speed improvements using our neural codes over using the original video frames; and (3) the ability to use standard tools, such as augmentations, within our framework to recreate the generic vision pipeline.

We investigate this by first determining how well the compressed tensors can be used to reconstruct the original videos in Section~\ref{sec:reconstruction} before discussing their utility on a variety of downstream tasks in Sections~\ref{sec:kinetics600}-\ref{sec:walkingtours}. We also investigate transfer of neural codecs if they are trained on another dataset.
Next, we investigate how well our trainable  transformations  model augmentations in Section~\ref{sec:equivariant}.
Finally, we discuss the speed of a classification model when used with neural codes in Section~\ref{sec:speed}.
We show more details in the supplementary material.

\subsection{Datasets}
We pre-train our models on one of two datasets: Kinetics600 \cite{kay2017kinetics,carreira2018short} and our internal dataset of recordings of tourists visiting different places. We named this dataset WalkingTours. We evaluate our models on three datasets: Kinetics600 (video classification), WalkingTours (hour long understanding), and COIN (framewise video classification) \cite{tang2019coin}. Further details are in the supplementary.

\subsection{Reconstructions}
\label{sec:reconstruction}
\begin{figure}[tb]
    \centering
    \tiny
    \includegraphics[width=0.62\linewidth,trim={00.3cm 0cm 0.3cm 0cm},clip]{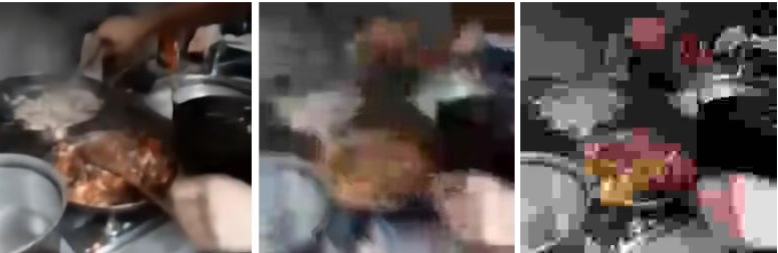}
    \caption{{\bf Neural vs MPEG vs JPEG codecs.} Our neural codecs (left) are better than MPEG (middle) and JPEG (right) at higher compression rates ($\approx$180 CR).
    }
    \label{app:fig:neural_vs_mpeg}
\end{figure}

We first investigate the quality of our neural codes for the task of reconstruction.
Note, however, that in our work the final performance on downstream tasks is more important than the reconstruction benchmark; in the spirit of~\cite{dubois2021lossy}.
Nonetheless, we would expect these numbers to be correlated with downstream performance. We can also use reconstruction  for introspection and debugging.
\pp{Baselines.} We compare our results to using a JPEG encoding. This is a commonly used compression scheme in video pipelines, and even though it ignores the time dimension, it offers quite good quality at low compression rates.
However, unlike the frame-based JPEG, our compressors have access to neighbouring information in order to learn better representations at potentially larger compression rates.
We use the OpenCV library \cite{opencv_library} to obtain the JPEG encodings for various compression levels. We also compare the quality of our compression mechanism to MPEG, which is better than JPEG at higher compression rates. 
\pp{Results.} We report the performance of our compression model when reconstructing the original videos in  Table~\ref{tab:reconstruction} and visualise reconstructions for varying compression levels in Figure~\ref{fig:reconstruction}.
We use standard reconstruction metrics for various levels of compression to demonstrate the reconstruction and compression trade-offs.
We can see that we can get low reconstruction error for large compression rates and 
that these reconstructions 
capture the higher level structure of the scene and are good quality. We also observe much higher degradation in quality for higher compression rates in JPEG and MPEG than our neural codec. Our visualizations in Figure~\ref{app:fig:neural_vs_mpeg} confirm the quantitative results in Table~\ref{tab:reconstruction} that our codec is better than MPEG at high compression rates. 
This demonstrates that our representations should be informative enough to use for downstream tasks.

\begin{table}[h]
    \centering
    \begin{tabular}{c ccccc} \\ \toprule
    & \multicolumn{3}{c}{Kinetics600} \\
          &  PSNR $\uparrow$ & SSIM $\uparrow$ & MAE $\downarrow$  \\ \midrule
         JPEG CR$\sim$30 & 36.4 & 94.1 & 0.013   \\
         JPEG CR$\sim$90 & 25.1 & 70.2 & 0.045  \\
         JPEG CR$\sim$180 & 22.5 & 63.1 & 0.057  \\ \midrule
         MPEG CR$\sim$30 & 33.2 & 89.6 & 0.034  \\
         MPEG CR$\sim$90 & 38.7 & 82.4 & 0.026 \\
         MPEG CR$\sim$180 & 23.7 & 67.3 & 0.054  \\ \midrule
         CR$\sim$30 & 38.6 & 97.6 & 0.008    \\
         CR$\sim$236 & 30.8 & 89.8 & 0.019   \\
         CR$\sim$384 & 30.0 & 88.4 & 0.019  \\
         CR$\sim$768 & 29.0 & 85.4 & 0.022  \\ \bottomrule
    \end{tabular}
    \caption{{\bf Reconstruction error for codecs.} We compare our approach to using JPEG and MPEG encodings. We report three standard reconstruction metrics (PSNR, SSIM, the mean absolute  error (MAE)) on Kinetics600 at different compression rates (CRs).
    For MAE, lower is better, for others, higher is better.}
    \label{tab:reconstruction}
\end{table}

\subsection{Video-Level Classification on Kinetics600}
\label{sec:kinetics600}

Next, we investigate directly using our neural codes for downstream tasks.
\pp{Baselines.} We compare to the baseline of just using the original RGB frames.
This is an upper bound of what we would expect to achieve.
In our case, instead of inputting a tensor of frames into the downstream network for classification, we input the compressed tensors directly.
To apply the S3D architecture to these compressed tensors we simply modify the stride and kernel shapes of the network (hyperparameters of the model). 
Note that such small modifications are common in the research on videos, e.g.~when using `space-to-depth' tricks.
\pp{Results.}
Results are reported in Table~\ref{tab:downstreamkinetics600}. 
As we can see, using 30x compression leads to a small ($\sim 1\%$ drop in performance) and we can even use on the order of 256x or 475x compression with only a $5\%$ drop in performance. In this research, we have simplified the pipeline, e.g.~we use fewer iterations than commonly used, hence, our results are below SOTA using the same model.

We also investigate how well the neural codes transfer: if we use neural codes trained on a different dataset, say WalkingTours, how well do they transfer to Kinetics600.
We find in Table~\ref{tab:downstreamkinetics600} that representations transfer between datasets; i.e. using a representation trained on WalkingTours leads to a similar result.

\begin{table}[h]
    \centering
    \begin{tabular}{cc c cc} \\ \toprule
    \quad & \quad & Evaluated on K600 \\
    \midrule
     \multicolumn{2}{c}{Trained on K600} & \quad & \multicolumn{2}{c}{Trained on WalkingTours} \\
          CR &  Top-1 $\uparrow$ &   \quad \quad \quad & CR & Top-1 $\uparrow$ \\ \midrule
         CR$\sim$1 & 73.1 & & CR$\sim$1 & 73.1  \\
         CR$\sim$30 & 72.2 & & CR$\sim$30 & 71.3  \\
         CR$\sim$475 & 68.2 & &  CR$\sim$256 & 68.4 \\ \bottomrule
    \end{tabular}
    \caption{{\bf Classification accuracy on Kinetics600.} 
    We report Top-1 accuracy on K600 when using neural codes trained one either K600 or WalkingTours. We experiment with different levels of compression (different compression rates (CRs)). CR$\sim$1 denotes original RGB frames (without compression).
    }
    \label{tab:downstreamkinetics600}
\end{table}

\subsection{Frame-level Classification on COIN}
\label{sec:coin}
We next investigate the performance of our neural codes on a different task: framewise prediction.
Unlike in the video classification task, this requires being able to predict localised information.
A representation that throws away too much information would struggle on this task.
Here we investigate the utility of our neural codes on this task.
\pp{Setup.} We use the same RGB baseline and setup as in the Kinetics600 case (Section~\ref{sec:kinetics600}).
As we found in Section~\ref{sec:kinetics600} that training the neural codes on a different dataset led to no loss of performance, we use our neural codes trained on either Kinetics600 or WalkingTours when training the downstream model.
\pp{Results.}
We report per frame accuracy in Table~\ref{tab:coin} for both datasets and different compression rates.
We compute means and standard deviations over the test set.
We find that, surprisingly there is virtually no loss of performance at a 30x compression rate and only minimal loss of performance at higher compression rates (e.g.~256x, 475x). 
Indeed, at 30x compression our model trained on Kinetics600 is even performing better than using the original RGB frames.
The choice of training dataset for a neural codec has minimal impact on the downstream performance.

\begin{table}[]
    \centering
    \begin{tabular}{ccc c ccc} \\ \toprule
        \quad & \quad & \quad & Evaluated on COIN \\
        \midrule
     \multicolumn{3}{c}{Trained on K600} & \quad & \multicolumn{3}{c}{Trained on WalkingTours} \\
          CR &  Top-1 $\uparrow$ &  Top-5 $\uparrow$ &   \quad \quad \quad & CR & Top-1 $\uparrow$ &  Top-5 $\uparrow$ \\ \midrule
         CR$\sim$1 & 44.3 $\pm$ 0.3 & 71.7 $\pm$ 0.1 & & CR$\sim$1 & 44.3 $\pm$ 0.3 & 71.7 $\pm$ 0.1  \\
         CR$\sim$30 & 45.5 $\pm$ 0.4 & 73.2 $\pm$ 3.7 & & CR$\sim$30 & 44.6 $\pm$ 0.3 & 71.9 $\pm$ 0.3  \\
         CR$\sim$475 & 41.7 $\pm$ 0.5 & 65.6 $\pm$ 0.5 &  & CR$\sim$256 & 42.7 $\pm$ 0.3 & 67.1 $\pm$ 0.3 \\ \bottomrule
    \end{tabular}
    \caption{{\bf Downstream classification accuracy on COIN.} We report Top-1 and Top-5 accuracy on COIN when using neural compression trained on either K600 or WalkingTours. We experiment with different levels of compression (different compression rates (CRs)). CR$\sim$1 denotes original RGB frames.}
    \label{tab:coin}
\end{table}

\subsection{Augmentations in the Compressed Space}
\label{sec:equivariant}
\begin{table}[]
    \centering
    \begin{tabular}{c c cccc} \\ \toprule
    & & \multicolumn{4}{c}{Num of temporal clips} \\
     & Crop Size  & 1 & 2 & 4 & 8 \\ \midrule
     224 central crop & 224 &  60.6 & 62.1 & 67.8 & 69.6 \\
     224 NN Crops (4 spatial crops)~\cite{patrick2021space} & 224   & 60.7 & 63.0 & 68.1 & 69.0 \\ \midrule
     Ours (2 spatial crops) & 224  & 61.6 & 64.1 & 69.1 & {\bf 70.1} \\
     Ours (3 spatial crops) & 224  & 61.3 & 63.9 & 68.9 & 69.6 \\
     Ours (4 spatial crops) & 224  & {\bf 61.9} & {\bf 64.4} & {\bf 69.3} & 69.6 \\ \midrule \midrule
     256 central crop & 256 & 60.8 & 62.4 & 68.2 & 68.9 \\ \midrule
     Ours (with flipping at train)& 256  & 61.7 & 64.4 & 68.5 & 70.0 \\ 
     Ours (with flipping at train and eval)& 256  & {\bf 62.9} & {\bf 65.2} & {\bf 69.0} & {\bf 70.2} \\ \bottomrule
     
    \end{tabular}
    \caption{{\bf Using our learnt network for augmentation.} We report Top-1 accuracy on K600. We experiment with different numbers of spatial crops at resolution $224\times224$ and with flipping at resolution $256\times256$. When using spatial crops, all models are trained using augmented spatial crops. 
    }
    \label{tab:equivariant}
\end{table}

\begin{figure}[h]
    \centering
    \tiny
    \begin{overpic}[width=0.32\linewidth,trim={0.3cm 0cm 0.3cm 0cm},clip]{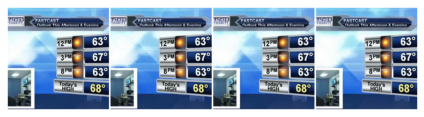}
    \put(-3,15){\makebox(0,0){\rotatebox{90}{Input}}}
    \end{overpic}
    \includegraphics[width=0.32\linewidth,trim={00.3cm 0cm 0.3cm 0cm},clip]{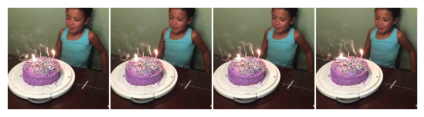}
    \includegraphics[width=0.32\linewidth,trim={0.3cm 0cm 0.3cm 0cm},clip]{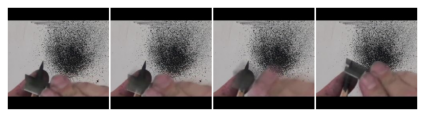}
    
    \vspace{-0.5em}
    
    \begin{overpic}[width=0.32\linewidth,trim={0.3cm 0cm 0.3cm 0cm},clip]{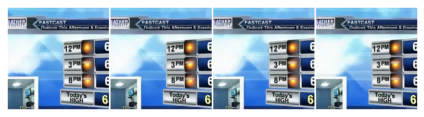}
    \put(-3,15){\makebox(0,0){\rotatebox{90}{Crop (1)}}}
    \end{overpic}
    \includegraphics[width=0.32\linewidth,trim={0.3cm 0cm 0.3cm 0cm},clip]{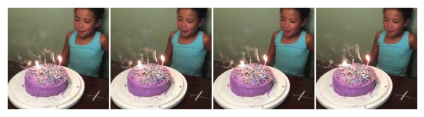}
    \includegraphics[width=0.32\linewidth,trim={0.3cm 0cm 0.3cm 0cm},clip]{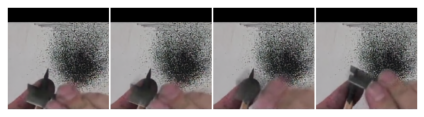}

    \vspace{-0.5em}
    
    \begin{overpic}[width=0.32\linewidth,trim={0.3cm 0cm 0.3cm 0cm},clip]{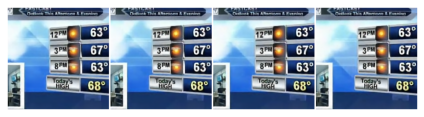}
    \put(-3,15){\makebox(0,0){\rotatebox{90}{Crop (2)}}}
    \end{overpic}
    \includegraphics[width=0.32\linewidth,trim={0.3cm 0cm 0.3cm 0cm},clip]{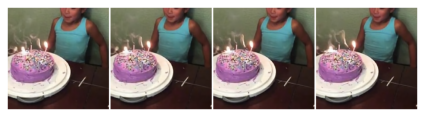}
    \includegraphics[width=0.32\linewidth,trim={0.3cm 0cm 0.3cm 0cm},clip]{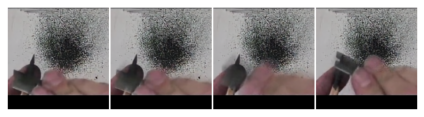}
    
    \vspace{-0.5em}
    
    \begin{overpic}[width=0.32\linewidth,trim={0.3cm 0cm 0.3cm 0cm},clip]{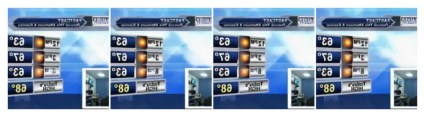}
    \put(-3,15){\makebox(0,0){\rotatebox{90}{Flip}}}
    \end{overpic}
    \includegraphics[width=0.32\linewidth,trim={0.3cm 0cm 0.3cm 0cm},clip]{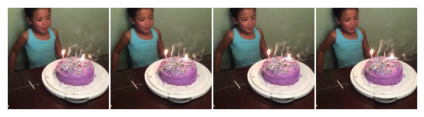}
    \includegraphics[width=0.32\linewidth,trim={0.3cm 0cm 0.3cm 0cm},clip]{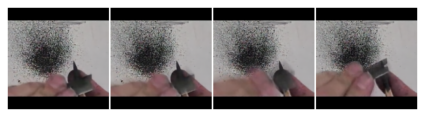}
    
    \caption{{\bf Learned transformations for augmentations.} The top row shows the original frames for three videos; the next two rows show these frames after applying our \augmentor for spatial cropping with two different bounding box inputs; and the bottom row shows them after applying our \augmentor for flipping. These results are obtained by applying our learned \augmentor to the corresponding compressed codes and decompressing.
    }
    \label{fig:equivariantresults}
\end{figure}

We also investigate whether we can harness standard techniques in a vision pipeline to push performance further.
A standard technique when achieving high quality results is to use spatial and temporal cropping at both training and evaluation time.
While our previous results assume that we can simply save various augmented versions of the dataset, that is not feasible for larger datasets.
\pp{Setup.} Here we investigate whether we can use an \augmentor to augment neural codes, as described in Section~\ref{sec:equivariant_representation}.
We train the augmentation network on Kinetics600, and use it to augment neural codes when training a downstream model.
In our experiments, we use the compression model trained on WalkingTours  with 30x compression rate.
We focus on spatial cropping and flipping --  standard augmentations used in video tasks to improve performance. 

For spatial cropping, we proceed as follows.
We take a video of size 256$\times$256 and randomly select a 224$\times$224 crop. This defines our bounding box which we use to learn the \augmentor. 
At train time, we randomly select a crop. 
At evaluation time, we pool over linearly spaced crops to obtain the final logits.
To investigate the utility of our approach, we investigate how we can improve performance at test time with additional, learned spatial crops (after training the downstream model with augmented crops). 

For flipping, we take a video of size 256$\times$256.
At train time we either flip or not.
At evaluation time, we pool over the original clip and its flipped version.
We evaluate whether flipping (at train, eval or both) improves performance.
\pp{Baselines.} 
For spatial cropping, we compare to two baselines.
First, we consider simply using 224$\times$224 central crops.
Second, we crop directly the compressed tensor by using nearest neighbours sampling according to the bounding box resized to that spatial resolution; this baseline is similar to the approach in \cite{patrick2021space}. 
For flipping, we compare performance to using a 256$\times$256 central crop.
\pp{Results.} We visualise our learned augmentations in Figure~\ref{fig:equivariantresults} for spatial cropping and flipping.
As can be seen, these augmentations closely match the ground truth transformation.
This observation is quantitatively confirmed by the SSIM scores (0.96) between the results obtained by directly cropping in the RGB-space and using our learned \augmentor. As a reference the same score between unchanged videos and doing crops in the RGB-space yields only 0.44 SSIM score, showing that the ground-truth and learned crops are indeed highly correlated. We draw similar conclusions about the other two augmentations.
Regarding downstream tasks, we report results in Table~\ref{tab:equivariant} for Kinetics600.
As we can see, learning to approximate augmentations improves over all baselines, demonstrating that our pipeline can leverage standard data augmentation techniques to push performance on downstream tasks.
An additional benefit of our learned approach is that we can model transformations that cannot be defined by a simple transformation over embeddings, e.g. flipping or changing the brightness. The appendix has more such examples (e.g.~saturation and rotation).

Note that the setup for these results differ from those in Table~\ref{tab:downstreamkinetics600}.
In Table~\ref{tab:downstreamkinetics600}, we augment the videos before creating codes by using random spatial cropping. 
Here we {\em only} use the learned augmentation at train and test time to augment the neural codes with transformed versions.
Using learned augmentations leads to a 2\% drop in performance while improving over the baselines. It is because we investigate augmentations after applying central crops; effectively reducing the space of possible combinations. This is mostly an engineering limitation as the existing pipelines do cropping before assembling data into batches.

Finally, we also show that we can train our \augmentor to do other transformations that are unnecessary for the classification tasks, but could potentially be useful for other problems. These results also show how universal our methodology is. Figures~\ref{fig:equivariantbrightness} and \ref{app:fig:equivariantbrightnessmultiple}, in the supplementary material, show the brightness transformations. Figure~\ref{fig:equivariantrotationsaturation} shows other two challenging transformations: rotations and changes in saturation. As we can see all these transformations are successfully learned. Note that, all these transformations are conducted in a latent space, and we decode them afterwards for the convenience.

\subsection{Long-Term Video Predictions}
\label{sec:walkingtours}
Here, we discuss our results on long-term video prediction using WalkingTours as the dataset. Towards this goal we have created a task {\em past-future} that does not require supervision.
{\em Past-future} provides a query to the network and asks the question if the short video clip (5 seconds) has been observed by the network at some point in the past. Sampled clips always come from the same video but have different time stamps.
We do the inference, at train and test times, in the causal setting, where the network only sees inputs seen in the past and has no access to the future frames. All the past frames form a compressed memory and are directly accessible to the network at the query time. 

Our downstream model differs from the models used above as it requires access to the memory. For that we use a transformer architecture that can effectively query the memory. We also experiment with different memories. Non-parametric {\em slot} refers to directly keeping all the past frames in the memory (but in the compressed form). We also use {\em LSTM} as the memory; it transforms a variable-length inputs into a fixed-length vector representation. Finally, {\em none} denotes no memory. 
We provide a more detailed explanation of the architecture and the task in the supplementary material. We have conducted experiments on 30 minutes long videos (training and inference). Our results give $99.6\%$ for {\em slot}, $78.2\%$ for {\em LSTM} and $52.9\%$ for {\em none} (which is equivalent to random chance). Using one-hour long videos, with our best configuration, leads to a drop in performance of $66\%$, demonstrating there are still gains to be made for long video understanding.

\subsection{Speed requirements}
\label{sec:speed}
Here, we investigate whether we can train architectures faster; as
our representations are smaller spatially and in total memory size.
To perform this comparison we run the forward pass $100$ times and report the mean and standard deviation of the time it takes on a Tesla V100.
We compare the speed when using a compression rate of 256 versus no compression rate.
Using compression rate of 256 leads to a forward pass that is on average $0.052 \pm 0.002$ seconds versus $0.089 \pm 0.001$ when using no compression.
Thus, we can achieve about 2x speed-up with minimal loss of accuracy. The supplementary material has more comprehensive study.

\section{Conclusions}

This work recreates a standard video pipeline using a neural codec.
After learning our neural codes, we can (1) use competitive, modern architectures and (2) apply data augmentation directly on the neural codes.
The benefits of using our compressed setup is that we can (1) reduce memory requirements; (2) improve training speed; and (3) generalise to minute or even hour long videos.
In order to investigate the feasibility of our approach on hour long videos, we introduced a dataset and task to explore this setting.

Our work is the first, to our knowledge, to demonstrate 
how to use  neural codes for a set of, potentially unknown, downstream tasks by leveraging standard video pipelines, including data augmentation in the form of learned transformations on the compressed space.
There are many avenues of future work to improve video understanding at scale.
Some directions include improving the compression, scaling to multi-day videos, and collating tasks and datasets for such long horizon timescales.
\pp{Acknowledgements.}
We thank Sander Dieleman for feedback on the paper.
We thank Meghana Thotakuri for her help with open sourcing this work.

\clearpage

 \setcounter{page}{1}

\clearpage

\appendix

\section{Overview}
We report additional results in Section~\ref{app:sec:results} including on the reconstruction task (Section~\ref{app:sec:reconstruction}) and when using augmentations for downstream tasks (Section~\ref{app:sec:equivariantresults}).
We visualise additional learned augmentations (e.g.~brightness, saturation and rotation) in  Section~\ref{app:sec:equivariantresults} and give additional visualisations of the cropping augmentation.
We also provide a video comparing our compressed representation to that of MPEG and visualising our learned augmentations.

We provide a more comprehensive study on the speed of various components of our setup in Section~\ref{app:sec:speed} and additional information about the datasets in Section~\ref{app:dataset}. Section~\ref{app:wtpipeline} has a more detailed exposition on WalkingTours -- our new dataset and task for handling very long video sequences.
Finally, we put together a more detailed description of our models in Section~\ref{app:model}, including how we apply S3D on our neural codes (Section~\ref{app:s3dapplied}) as well as information about the other architectures, the precise hyperparameter sweeps we consider (Section~\ref{app:sweeps}) and training details (Section~\ref{app:training}).

\section{Additional Results}
\label{app:sec:results}

\subsection{Reconstructions at Various Compression Rates}
\label{app:sec:reconstruction}
We expand the results from the main paper with reconstruction performance at various compression rates in Table~\ref{app:tab:reconstruction}. 
As before, we can see that at the same compression rate, the reconstruction performance using JPEG encodings degrades much more quickly than when using our neural codes; for instance, we can compress inputs four times more using our neural compression than the JPEG compression.

\begin{table}[h]
    \centering
    \begin{tabular}{c ccccc} \\ \toprule
    & \multicolumn{3}{c}{Kinetics600}  \\
    &  PSNR $\uparrow$ & SSIM $\uparrow$ & MAE $\downarrow$  \\ \midrule
         JPEG CR$\sim$30 & 36.4 & 94.1 & 0.013   \\
         JPEG CR$\sim$90 & 25.1 & 70.2 & 0.045  \\
         JPEG CR$\sim$180 & 22.5 & 63.1 & 0.057  \\ \midrule
         MPEG CR$\sim$30 & 33.2 & 89.6 & 0.034  \\
         MPEG CR$\sim$90 & 38.7 & 82.4 & 0.026 \\
         MPEG CR$\sim$180 & 23.7 & 67.3 & 0.054  \\ \midrule
         CR$\sim$30 & 38.6 & 97.6 & 0.008    \\
        CR$\sim$90 & 34.8 & 94.8 & 0.013  \\
         CR$\sim$180 & 32.6 & 92.3 & 0.016  \\
         CR$\sim$236 & 30.8 & 89.8 & 0.019   \\
         CR$\sim$384 & 30.0 & 88.4 & 0.019  \\
         CR$\sim$768 & 29.0 & 85.4 & 0.022  \\ \bottomrule
    \end{tabular}
    \caption{{\bf Reconstruction error for neural codes.} Additional results. We compare our approach to using JPEG encodings of the frames. We report three standard reconstruction metrics (PSNR, SSIM and the mean absolute  error (MAE)) for training on Kinetics600 at different compression rates (CRs). For MAE, lower is better whereas for PSNR and SSIM~\cite{SSIM_reference}, higher is better. }
    \label{app:tab:reconstruction}
\end{table}

\subsection{Compression Trade-offs}
In addition to our main results, we have also experimented with various hyper-parameters, showing different trade-offs.
\pp{Space- vs Time-compression.} Alternatively to downsampling spatial resolution (space-compression) we could also downsample temporal resolution (time-compression). For that purpose, we use temporal striding in convolutional kernels. To keep the same compression rate, CR 236, we use the following setup. In space-compression, we compress 32-frames long video with the spatial resolution $256-by-256$ into a tensor of shape:  $32 \times 16 \times 16 \times 2$ (time x width x height x number of vocabularies). In time-compression, we compress the same video into a tensor of shape: $4 \times 32 \times 32 \times 4 $ (time x width x height x number of vocabularies). Space-compression yields the SSIM score $89.3$, whereas time-compression $89.8$. Both results are comparable though there are small qualitative differences in the decoded videos for individual videos.
\pp{Codebook size vs number of codebooks.} VQ-VAE~\cite{oord2017neural} uses a fixed number of codes in a single codebook. Here, for a fixed compression ratio, we can either increase number of codes in a single codebook or use different codebooks. Note that, if we decide to increase the number of codebooks twice, we are free to increase the number of codes quadratically as 
\[
c_r = \frac{I_T I_H I_W * 3 * \log_2 256}{T_T T_H  T_W  (2 T_C) \log_2 K} = \frac{I_T I_H I_W * 3 * \log_2 256}{T_T T_H  T_W  T_C \log_2 K^2}
\]
where $c_r$ is the compression rate, $I_H, I_W$ are spatial resolutions for each frame, $I_T$ is the number of frames in a single video, $T_T T_H, T_W$ denotes the shape of the compressed spatio-temporal tensor, $T_C$ is the number of codebooks and $K$ is the number of codes in each codebook (we need to store indices to these codes). Even though compression rates are the same, we found that it is better to increase the number of codebooks at the cost of fewer codes per codebook. That is, a single codebook with 65k codes gives worse results than two codebooks, with 256 codes each. The corresponding SSIM scores are 86 and 88.

\subsection{Using Compressed Representations versus Reconstructions}
In the paper, we train downstream tasks directly using the neural codes.
However, at the cost of significant speed loses as shown in Section~\ref{app:sec:speed}, we could take the neural codes, pass them through the generator $c^{-1}$ and obtain the reconstructed images. This generation / reconstruction process is akin to the decompression process in standard (non-neural) codecs.
We could then use these reconstructed images to train the downstream tasks.
In Table~\ref{app:tab:generatedresults}, we compare the performance between using the neural codes and reconstructed images on Kinetics600.
We find that there is a small drop in performance between using the neural codes and reconstructed images.
This demonstrates that if the representation better captures the full image, we would expect improvements in performance (i.e.~the drop in performance is not from the compression itself but from the quality of the learned representation).

\begin{table}[h]
    \centering
    \begin{tabular}{cc c cc} \\ \toprule
     \multicolumn{2}{c}{K600} & \quad & \multicolumn{2}{c}{WalkingTours} \\
          CR &  Top-1 $\uparrow$ &   \quad \quad \quad & CR & Top-1 $\uparrow$ \\ \midrule
         \multicolumn{5}{l}{\bf Original images} \\
         CR$\sim$1 & 73.1 & & CR$\sim$1 & 73.1  \\ \midrule
         \multicolumn{5}{l}{\bf Reconstructed images} \\
         CR$\sim$30 & 71.2 & & CR$\sim$30 & 72.8  \\
         CR$\sim$475 & 69.0 & & CR$\sim$256 & 71.4  \\ \midrule
         \multicolumn{5}{l}{\bf Neural codes} \\
         CR$\sim$30 & 72.2 & & CR$\sim$30 & 71.3  \\
         CR$\sim$475 & 68.2 & & CR$\sim$256 & 68.4  \\ \bottomrule
    \end{tabular}
    \caption{{\bf Downstream classification accuracy on Kinetics600.} Additional results. We compare using the neural codes directly versus using the reconstructed images. We report Top-1 accuracy on K600 when using neural codes trained on either K600 or WalkingTours. We experiment with different levels of compression (different compression rates (CRs)). CR$\sim$1 denotes the upper bound of using the original RGB frames. Using the neural codes as opposed to the reconstructed images leads to a minor drop in performance ($~1\%$), demonstrating that improving the quality of the representation would directly improve performance.} 
    \label{app:tab:generatedresults}
\end{table}

\subsection{Transferability of Augmentations}
Here, we consider whether the learned augmentations are transferable. That is, can we train the \compressor and \augmentor on one dataset and evaluate it on another? We compare the flipping and cropping transformations when the \compressor and \augmentor are trained on WalkingTours or Kinetics600.
Note that in the main paper, the \compressor was trained on WalkingTours and the \augmentor on Kinetics600.
As can be seen, using either the flipping or cropping augmentation, we improve over the baseline setup that does not use learnt augmentations.
This is valid even if the \compressor and \augmentor are trained on different datasets than the classification model (which is trained on Kinetics600). 
Pre-training both the \compressor and \augmentor on the same dataset as the classification network sometimes improves performance, but the difference between setups is marginal.
We did experiment with augmentations at larger compression rates but found that the results did not improve over the baselines; future work should explore how to learn augmentations at larger compression rates.

\begin{table}[]
    \centering
    \begin{tabular}{c c cccc} \\ \toprule
    & & \multicolumn{4}{c}{Num of temporal clips} \\
     & Crop Size  & 1 & 2 & 4 & 8 \\ \midrule
     224 central crop & 224 &  60.6 & 62.1 & 67.8 & 69.6 \\
     224 NN Crops (4 spatial crops)~\cite{patrick2021space} & 224   & 60.7 & 63.0 & 68.1 & 69.0 \\
     256 central crop & 256 & 60.8 & 62.4 & 68.2 & 68.9 \\ \midrule \midrule
     {\bf C: WalkingTours, A: Kinetics600} \\
     Ours (2 spatial crops) & 224  & 61.6 & 64.1 & 69.1 & {\bf 70.1} \\
     Ours (3 spatial crops) & 224  & 61.3 & 63.9 & 68.9 & 69.6 \\
     Ours (4 spatial crops) & 224  & {\bf 61.9} & {\bf 64.4} & {\bf 69.3} & 69.6 \\ \midrule
     Ours (with flipping at train)& 256  & 61.7 & 64.4 & 68.5 & 70.0 \\ 
     Ours (with flipping at train and eval)& 256  & {\bf 62.9} & {\bf 65.2} & {\bf 69.0} & {\bf 70.2} \\ \midrule \midrule
     {\bf C: WalkingTours, A: WalkingTours} \\
     Ours (2 spatial crops) & 224  & 61.2 & 64.1 & 69.2 & 69.2 \\
     Ours (3 spatial crops) & 224  & {\bf 61.5} & 64.7 & {\bf 69.5} & 69.5 \\
     Ours (4 spatial crops) & 224  & {\bf 61.5} & {\bf 64.8} & 68.9 & 69.5 \\ \midrule
    Ours (with flipping at train)& 256 & 61.8 & 63.9 & 68.3 & 69.5  \\ 
     Ours (with flipping at train and eval)& 256  & {\bf 62.6} & {\bf 65.1} & {\bf 69.3} & {\bf 70.4}  \\ \midrule \midrule
     {\bf C: Kinetics600, A: Kinetics600} \\
     Ours (2 spatial crops) & 224 & 61.3 & 62.9 & {\bf 68.3} & 69.3  \\
     Ours (3 spatial crops) & 224 & 61.3 & 63.2 & {\bf 68.3} & 69.2 \\
     Ours (4 spatial crops) & 224 & {\bf 61.6} & {\bf 63.6} & 68.2 & 69.3 \\ \midrule
     Ours (with flipping at train)& 256 & 62.9 & 64.1 & {\bf 70.1} & {\bf 70.8}  \\ 
     Ours (with flipping at train and eval)& 256  & {\bf 63.5} & {\bf 64.7} & 70.0 & 70.4 \\ \bottomrule
     
    \end{tabular}
    \caption{{\bf Using our learnt network for augmentation.} We report Top-1 accuracy on K600. We compare training the \compressor ({\bf C}) and \augmentor ({\bf A}) on the same or different datasets than what the classification network is trained on. For each group, we bold the best setup that improves over the baseline setup (which uses no learned augmentations). As can be seen, we can train the \augmentor on WalkingTours, Kinetics600, or a combination thereof and still improve on the original \compressor. }
    \label{app:tab:equivariant}
\end{table}

\subsection{Extra Augmentation Visualisations}
\label{app:sec:equivariantresults}

\paragraph{Cropping augmentation.}
We give additional visualisations of the cropping transformation in Figure~\ref{app:fig:croppingmultiple}. For a single image, we visualise four crops when that crop is applied (a) to the original image, (b) to the reconstructed image using our encoder-decoder network and (c) using our \augmentor applied to the neural codes.
As can be seen, there is minimal difference between the crops using our \augmentor and the original image.

\begin{figure}
    \centering
    \tiny
    \begin{overpic}[width=0.32\linewidth,trim={0.5cm 3cm 0.5cm 3cm},clip]{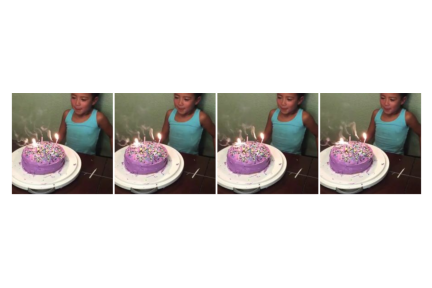}
     \put(-3,15){\makebox(0,0){\rotatebox{90}{Input}}}
    \end{overpic}
    \includegraphics[width=0.32\linewidth,trim={0.5cm 3cm 0.5cm 3cm},clip]{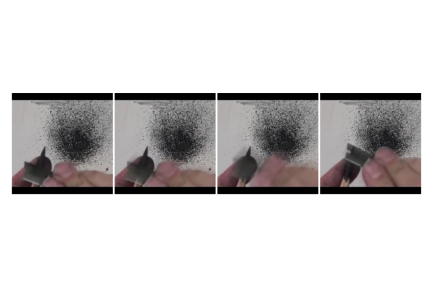}
    \includegraphics[width=0.32\linewidth,trim={0.5cm 3cm 0.5cm 3cm},clip]{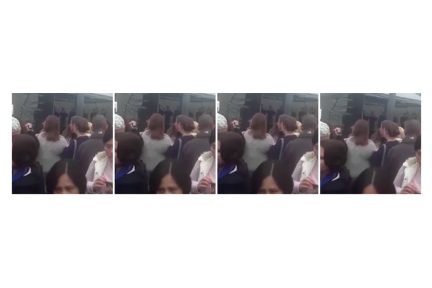}
    
    \vspace{-0.5em}
    
    \begin{overpic}[width=0.32\linewidth,trim={0.5cm 3cm 0.5cm 3cm},clip]{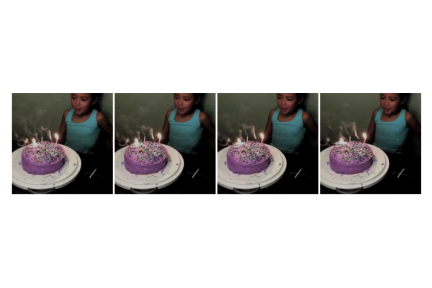}
     \put(-3,15){\makebox(0,0){\rotatebox{90}{Bright (1)}}}
    \end{overpic}
    \includegraphics[width=0.32\linewidth,trim={0.5cm 3cm 0.5cm 3cm},clip]{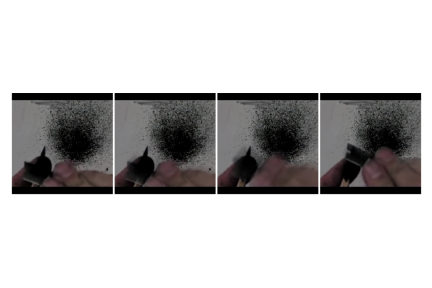}
    \includegraphics[width=0.32\linewidth,trim={0.5cm 3cm 0.5cm 3cm},clip]{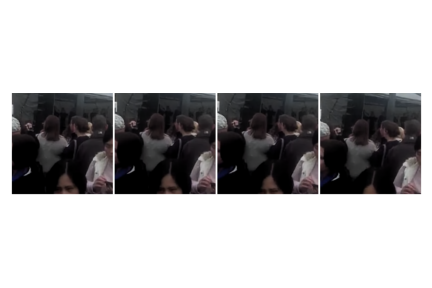}

    \vspace{-0.5em}
    
    \begin{overpic}[width=0.32\linewidth,trim={0.3cm 0cm 0.3cm 0cm},clip]{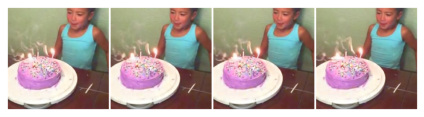}
    \put(-3,15){\makebox(0,0){\rotatebox{90}{Bright (2)}}}
    \end{overpic}
    \includegraphics[width=0.32\linewidth,trim={0.3cm 0cm 0.3cm 0cm},clip]{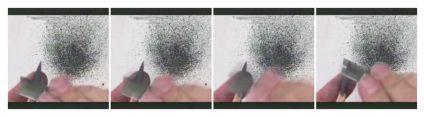}
    \includegraphics[width=0.32\linewidth,trim={0.3cm 0cm 0.3cm 0cm},clip]{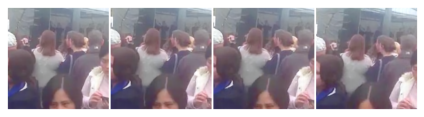}
    
    \caption{{\bf Learned augmentation: Brightness.} The top row shows the original frames for three videos; the bottom two rows show these frames after applying our equivariant network for brightness at two extremes.}
    \label{fig:equivariantbrightness}
\end{figure}

\begin{figure}
    \centering
    \begin{overpic}[width=\linewidth,trim={0cm 6cm 0cm 6cm},clip]{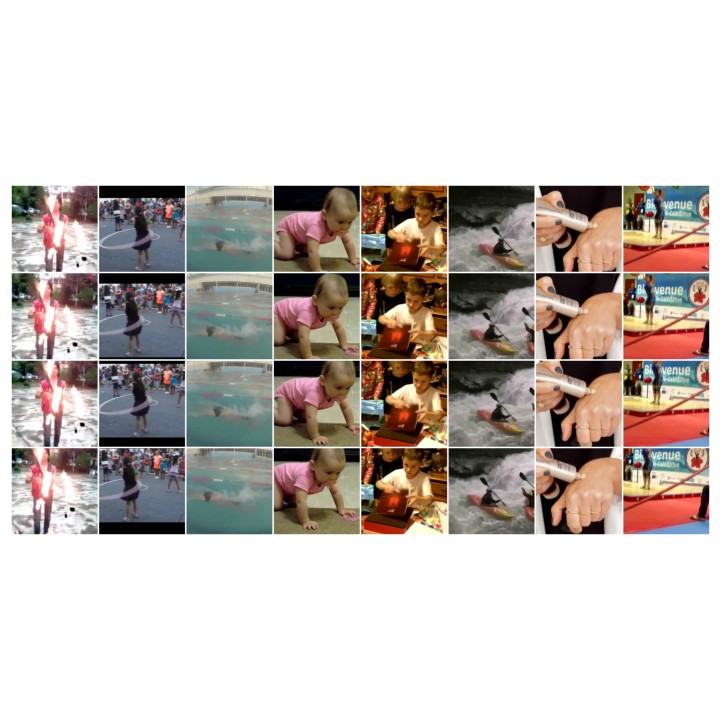}
    \put(0,26){\makebox(0,0){\rotatebox{90}{Original Images}}}
    \end{overpic}
    
    \begin{overpic}[width=\linewidth,trim={0cm 6cm 0cm 6cm},clip]{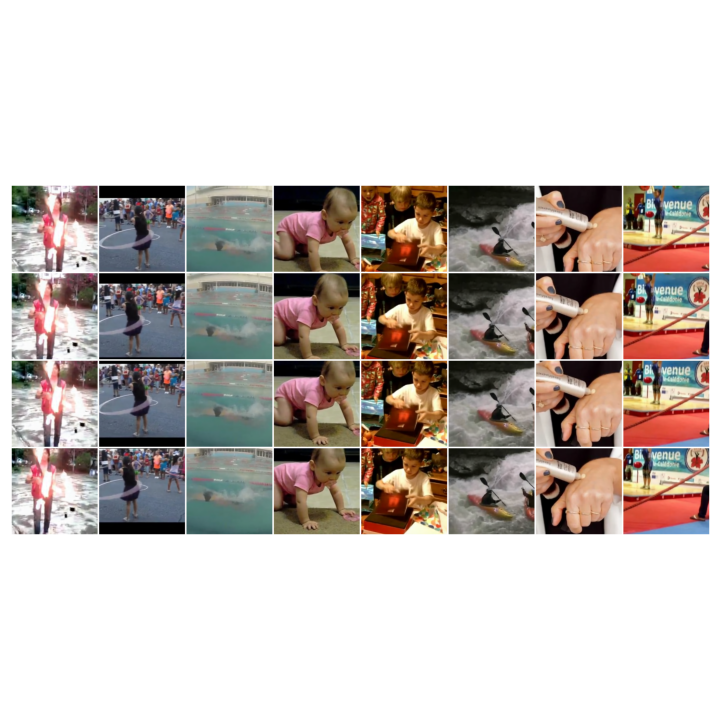}
    \put(0,26){\makebox(0,0){\rotatebox{90}{Reconstructed Images}}}
        \end{overpic}
    
    \begin{overpic}[width=\linewidth,trim={0cm 6cm 0cm 6cm},clip]{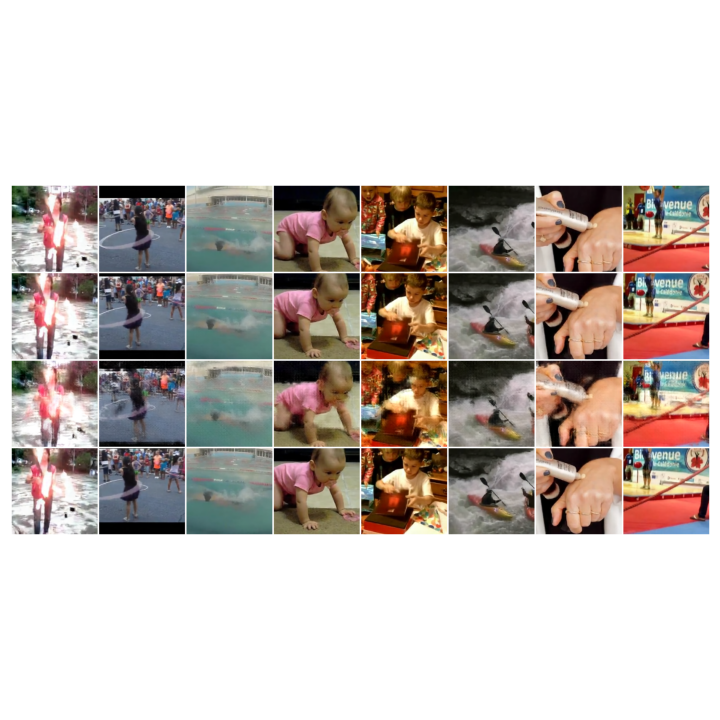}
    \put(0,26){\makebox(0,0){\rotatebox{90}{\augmentor on Neural Codes}}}
    \end{overpic}
    \caption{{\bf Learned augmentation: Cropping.} }
    \label{app:fig:croppingmultiple}
\end{figure}

\paragraph{Brightness augmentation.}
In this work, we mainly focus on learning transformations of the latent space for the purpose of augmentations, and thus, we focus on the most common ones. However, we can also learn other transformations such as changes in brightness, as shown in Figure~\ref{fig:equivariantbrightness} for three videos at two brightness extremes.
We have not found this transformation to improve performance of the classification models, so we show it here mainly to illustrate how our method is general. Note that, unlike when {\em cropping}, we cannot apply such a transformation by just manipulating neural codes (as in~\cite{patrick2021space}).

\begin{figure}
    \centering
    \begin{overpic}[width=\linewidth,trim={0cm 6cm 0cm 6cm},clip]{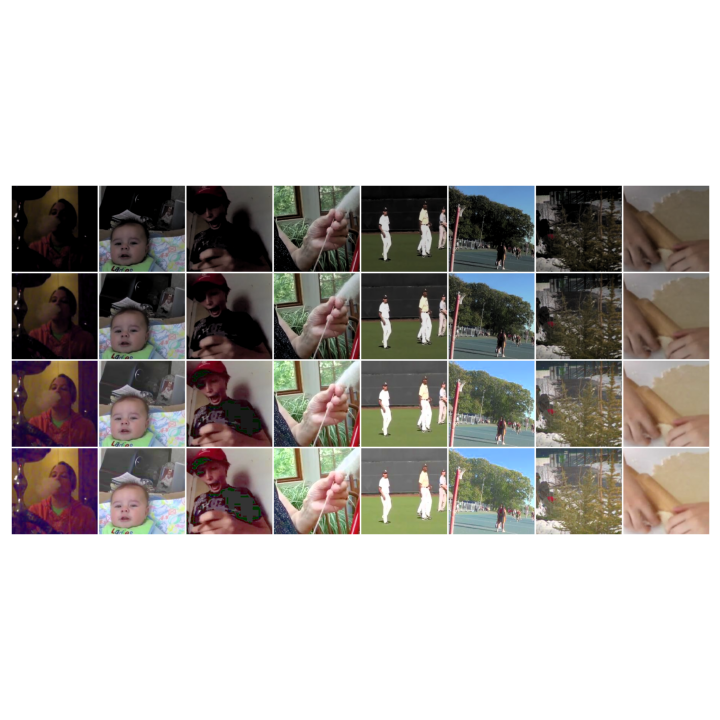}
    \put(0,26){\makebox(0,0){\rotatebox{90}{Original Images}}}
    \end{overpic}
    
    \begin{overpic}[width=\linewidth,trim={0cm 6cm 0cm 6cm},clip]{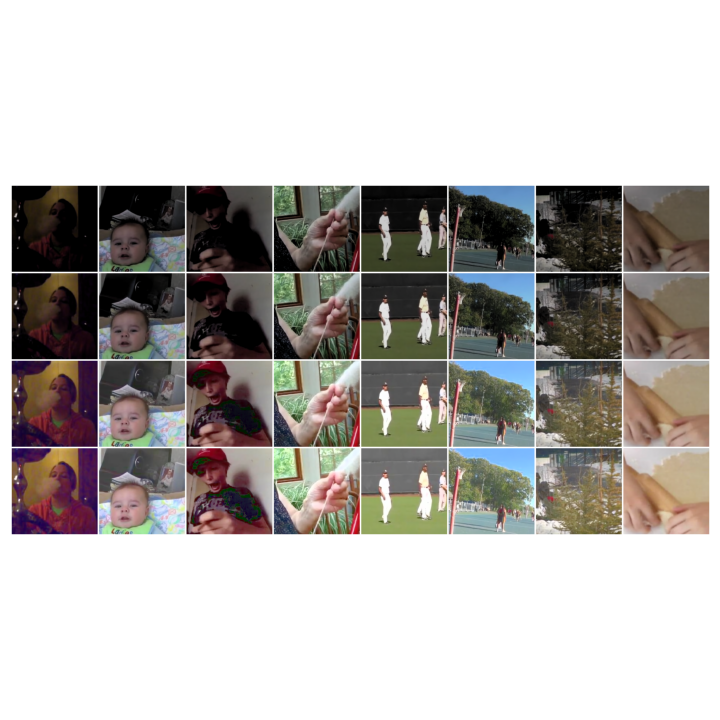}
    \put(0,26){\makebox(0,0){\rotatebox{90}{Reconstructed Images}}}
        \end{overpic}
    
    \begin{overpic}[width=\linewidth,trim={0cm 6cm 0cm 6cm},clip]{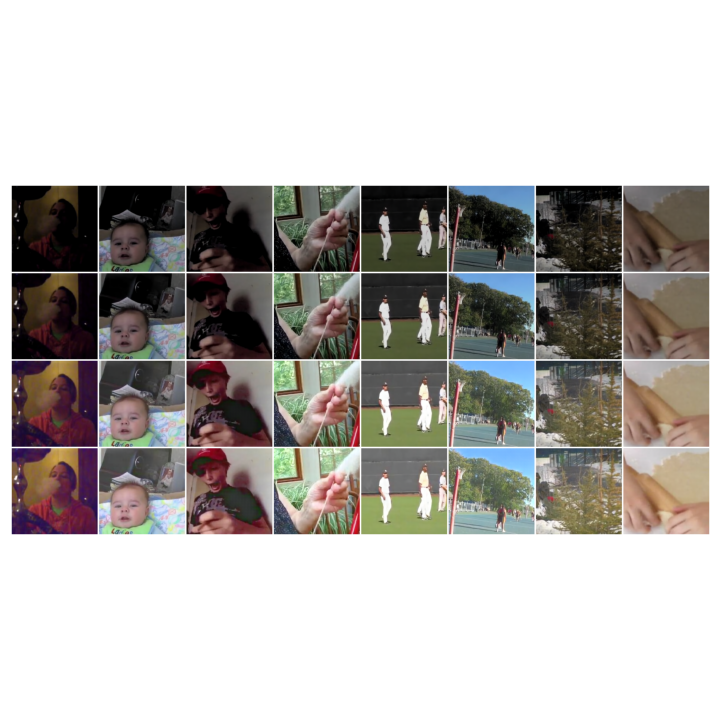}
    \put(0,26){\makebox(0,0){\rotatebox{90}{\augmentor on Neural Codes}}}
    \end{overpic}
    \caption{{\bf Learned augmentation: Brightness.} }
    \label{app:fig:equivariantbrightnessmultiple}
\end{figure}

Figure~\ref{app:fig:equivariantbrightnessmultiple} compares the results for multiple videos.
For a single image, four different amounts of brightness are applied.
As in Figure~\ref{app:fig:croppingmultiple}, we visualise results when the transformation is applied to (a) the original image, (b) the reconstructed image using our encoder-decoder network and (c) using our \augmentor applied to the neural codes.
Again, there is minimal difference between the results using our \augmentor and the original image.

\begin{figure}
    \centering
        \includegraphics[width=0.32\linewidth]{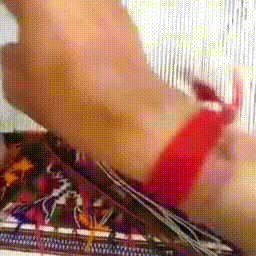}
    \includegraphics[width=0.32\linewidth]{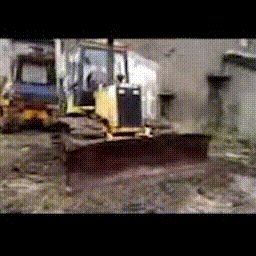}
    
    \vspace{-0.5em}
    
    \includegraphics[width=0.32\linewidth]{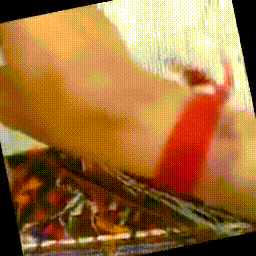}
    \includegraphics[width=0.32\linewidth]{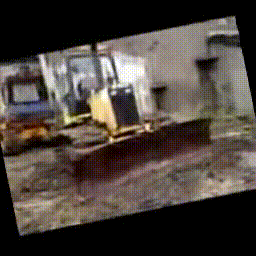}
    
    \vspace{-0.5em}
    
    \includegraphics[width=0.32\linewidth]{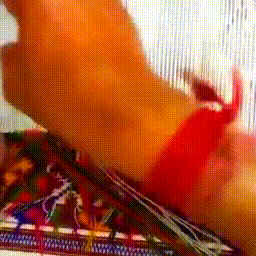}
    \includegraphics[width=0.32\linewidth]{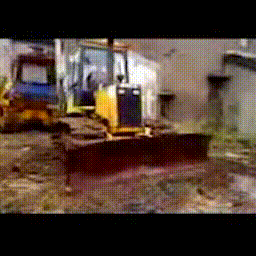}

    \caption{{\bf Learned augmentation: Rotations and Saturation.} Here, we show other, more challenging transformations. The top row presents the original video frames, middle row shows rotations whereas the bottom row saturation.}
    \label{fig:equivariantrotationsaturation}
\end{figure}

\paragraph{Other challenging transformations.}
 Figure~\ref{fig:equivariantrotationsaturation} shows other challenging transformations: rotations and changes in saturation. We see that the \augmentor can successfully learn such transformations.

\begin{figure}
    \centering
        \includegraphics[width=0.32\linewidth]{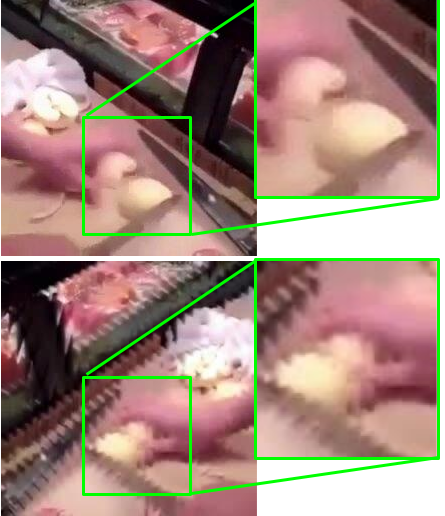}
        \put(10,118){\small \bf  \color{black} Original}
        \put(10,50){\small  \bf \color{black} Naive Flip}
    \caption{Naively flipping neural codes leads to strong artifacts after decoding.}
    \label{fig:naiveflips}
\end{figure}

\paragraph{Naive flips.}
Finally, many transformations operating in the compressed space cannot be easily constructed. For instance, to flip frames in the video, we cannot flip the codes as every individual code, which corresponds to a spatial region, needs to also be flipped. Naively performing this operation leads to problematic artifacts after decoding as shown in Figure~\ref{fig:naiveflips}.

\clearpage

\subsection{Time Measurements}
\label{app:sec:speed}
The compressed vision pipeline not only benefits from memory reduction, but also offers some speed gains. For instance, at compression rates of at least 256, we observe about $58\%$ reduction in the inference time compared to using the RGB-valued inputs (assuming we only measure the speed cost of the classification network). We do not observe speed gains at lower compression rates, such as 30. 

Table~\ref{tab:speed} gives detailed time measurements for the video inputs with 32 frames. More specifically, we measure the encoding and decoding times of our \compressor as well as the augmentation time with our \augmentor. We also show the time measurements of the \augmentor together with the inference of the Kinetics model (augment \& infer in the table). This is the most typical setting in our pipeline. This is because encoding can be done once in practice to store intermediary representation. Similarly, decoding is also not required in our pipeline, as we operate directly on the neural codes. This contrasts with standard vision pipelines that require decoding, and as we can see in~Table~\ref{tab:speed}, neural decoders incur a significant cost to the overall speed. Even JPEG decompression of a video with 32 frames, for the 30x compression rate and batch size 1  takes about 0.022s, which is slower than working directly with neural codes.
Thus, it is more efficient to avoid this step, and to perform the required computation in the compressed space. Note that decoders are often implemented using larger networks than encoders. In contrast, our \augmentor is a relatively shallow network, described in Section~\ref{app:augmentor}; this makes the whole pipeline efficient.

\begin{table}[]
    \centering
    \begin{tabular}{l c c c} \\ \toprule
     \textbf{Measurement} & \textbf{BS} & \textbf{CR} &  \textbf{Time} \\ \midrule
     Inference (rgb) & 8 & 1 & 0.089  \\
     Inference (codes) & 8 & 256 & 0.052 \\ \midrule
     Encoding & 8 & 256 & 1.894 \\
     Decoding & 8 & 256 & 3.076 \\ 
     Augmentation & 8 & 256 & 0.078  \\
     Augment \& infer & 8 & 256 & 0.129 \\
     \midrule
     Encoding & 1 & 30 & 0.816 \\
     Decoding & 1 & 30 & 1.328 \\
     Augmentation & 1 & 30 & 0.080  \\
     Augment \& infer & 1 & 30 & 0.102 \\
     \bottomrule
    \end{tabular}
    \caption{{\bf Speed measurements.} We report the inference time of the classification network, the encoding / decoding time of our \compressor, the augmentation time of our \augmentor, and the overall time of the combined augmentation and inference (augment \& infer). We report, in seconds, time averaged over 100 runs. The standard deviation is marginal in all cases. We use Tesla V100. We use batch size 8 for RGB and compression rate 256  to better utilize vectorized computations, and batch size 1 for compression rate 30 as otherwise we run into memory issues (column BS for batch size). CR denotes different compression rates.}
    \label{tab:speed}
\end{table}

\subsection{Comparison to Other Methods}

Here, we further discuss how our pipeline compares with other methods. 
We summarize these findings in Table~\ref{tab:comparisonothermethods}.
Other methods~\cite{chen2021fast,alizadeh2019compressed,shou2019dmc,li2020slow,wu2018compressed} operate directly on encoded frames (e.g.~I-Frames, P-Frames, and/or residuals) or may use additional motion vectors that are either learned or obtained from an MPEG representation.
That often requires devising new architectures; \cite{li2020slow} needs to add extra lateral connections that fuse two pathways at different frame-rates, and~\cite{shou2019dmc} trains a flow together with the downstream task.

While \cite{chen2021fast,wu2018compressed} use standard architectures (e.g.~ YOLO~\cite{redmon2016you} or ResNets~\cite{he2016deep}), they require a large amount of engineering complexity in manipulating the MPEG codes into an appropriate input modality (\cite{chen2021fast} partially decode the MPEG codes in order to perform pixel-level predictions), whereas we directly use our compressed codes with {\em no} further data engineering.
Moreover,~\cite{wu2018compressed} applies different models to different compressed representations yielding a codec-specific architecture.

Our compressed vision setup uses standard video architectures {\em without} requiring the development of an entirely new pipeline and we operate {\em directly} on the compressed codes with no further data engineering, and in a task-agnostic way. We also show how to perform augmentations directly in the compressed space.
Moreover, in Section~\ref{sec:reconstruction}, we find that our compression method performs better than MPEG at different compression rates; thus showing the need for trainable codecs. 
As our work is {\em not} MPEG specific, it could potentially benefit from better compressed representations.

\begin{table}[]
    \centering
    \begin{tabular}{lccc}
         Method \quad &\quad no MPEG \quad & no flow \quad & standard pipelines  \\ \toprule
         Alizadeh et al.~\cite{alizadeh2019compressed} & \xmark & \xmark & \xmark \\
         DMC-Net~\cite{shou2019dmc} & \xmark & \xmark & \xmark \\
         Li et al.~\cite{li2020slow} & \xmark & \xmark & \xmark \\
         Wu et al.~\cite{wu2018compressed} & \xmark & \checkmark & \xmark \\
         Chen et al.~\cite{chen2021fast} & \xmark & \checkmark & \checkmark \\
         Ours & \checkmark & \checkmark & \checkmark \\ \bottomrule
    \end{tabular}
    \caption{Comparison of our pipeline to other methods. We compare whether each method uses an MPEG style codec (e.g.~I-Frames or Blocks from that representation), a flow (optical flow, motion vectors, or their approximations), or whether the method leverages standard video pipelines (existing popular architectures and augmentations). Some methods operate on MPEG style representations. Thus they require new architectures, training schemes, or data engineering. In contrast, our pipeline can directly be used with the existing video architectures, and we can train the corresponding augmentations.
    Finally, other approaches rely on an MPEG style representation that lead to worse representation at higher compression rates to ours as demonstrated in Section \ref{sec:reconstruction}.
    }
    \label{tab:comparisonothermethods} 
\end{table}

\section{Datasets}
\label{app:dataset}
Here, we provide more information about the datasets.
\paragraph{Kinetics600} consists of short video clips downloaded from YouTube. The task is to predict which action corresponds to a given video, which is formalized as a classification problem (out of 600 classes) given the whole video.
However, these clips are up to 10 second long, and most video models are trained on about 2 second long clips at 25fps. Kinetics 600 has about 400k video clips for training purpose, and 100k video clips for evaluation, with less than 60 days of total video time.  Nonetheless, it is a popular benchmark in the research community, and even though our compressed vision framework is not essential, it can still make the whole training time much more efficient.
\paragraph{COIN} consists of relatively long video clips of the order of a couple of minutes; on average 2 minutes and 36 seconds. The dataset has 476 hours of total video time.
These are instructional videos with annotations per frame describing the task being shown in the frame. It has about 12k video clips.
We evaluate our approach on its ability to perform per-frame annotations as opposed to per-video classification as in Kinetics600. 
\paragraph{WalkingTours.} As we could not find any dataset containing sufficiently long videos, we decided to collect our own dataset as a proof of concept showing the utility of our proposed pipeline. The dataset consists of one-hour long videos of tourists walking in different places, and it poses various challenges for sampling and processing very long video sequences. These include high memory requirements for storage, high bandwidth, distributed and asynchronous requirements for sampling data and sending them to device, and high memory consumption on a device like GPU due to processing very large volumes of data points. Before even defining the right task on long video understanding, first these challenges above should be addressed.

The dataset has about 18k videos, with 1815 videos used for validation and the same number for a held-out test; the rest is the training set.
The videos range in length from 18 minutes to ten hours; on average 40 minutes. In total, the dataset amounts to around 500 days worth of accumulated video time.
It is significantly more than the number of days of accumulated video time in the latest Kinetics dataset~\cite{kay2017kinetics} and other egocentric datasets~\cite{Damen2018EPICKITCHENS,damen2020rescaling,grauman2021ego4d}. 
However, WalkingTours has no human annotations, and currently it is only compatible with self-supervised or unsupervised training or evaluation.

\section{Walking Tours: Task and Components}
\label{app:wtpipeline}
\begin{figure*}[t]
\begin{center}
\begin{tabular}{c@{\ }c}
\hspace{-1.0cm} \includegraphics[width=0.7\linewidth]{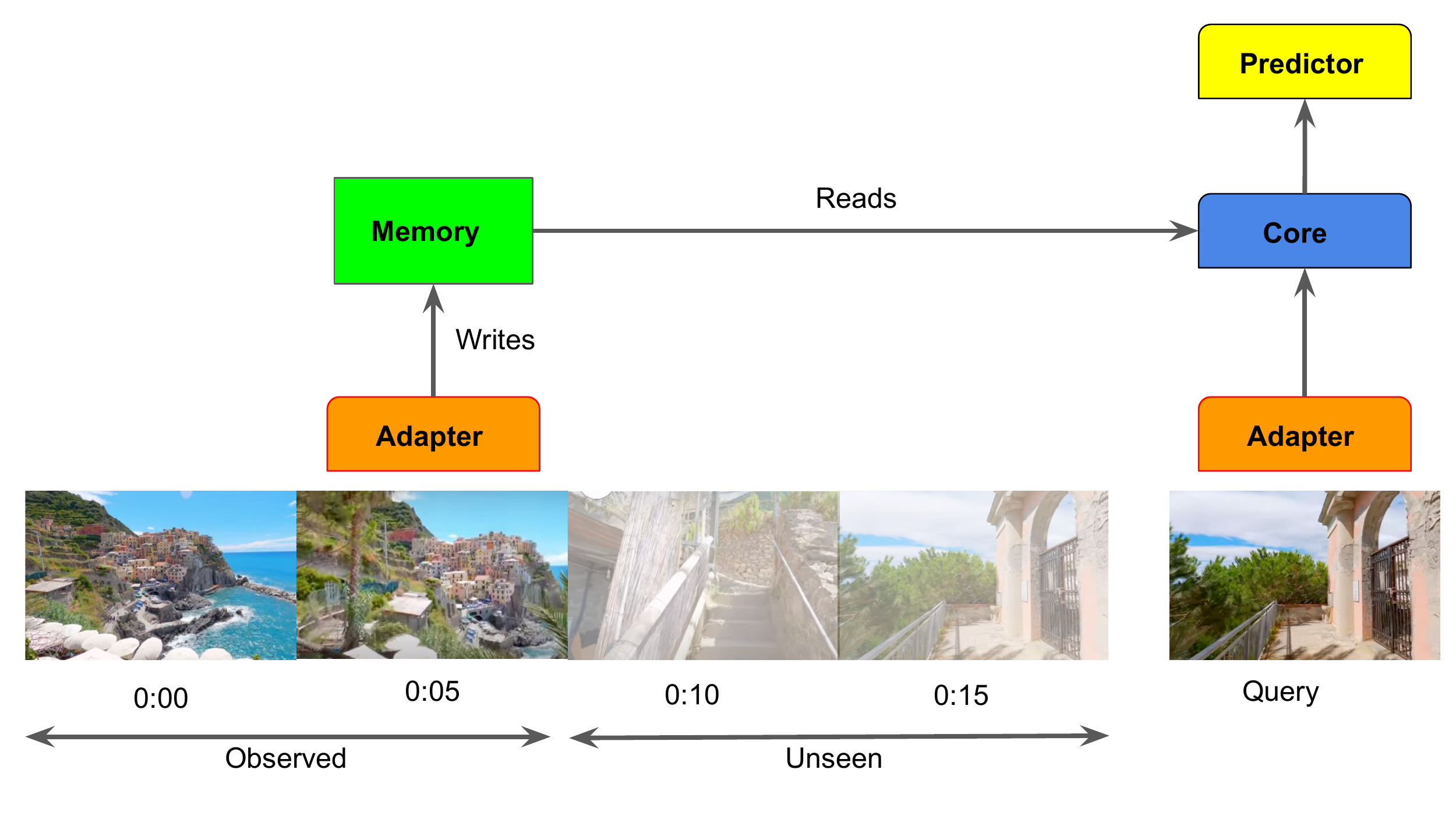} 
\end{tabular}
\end{center}
\caption{
Our online learning framework. We use the same colours to indicate weights sharing. Here, the query comes from the \textit{future}.
}
\label{fig:online_learning_tasks}
\end{figure*}

\paragraph{Description.}
WalkingTours is a dataset of very long, even one-hour long, videos. In the current form, it is also a purely visual dataset, i.e. there are no extra annotations associated with it; hence, models can only be trained and evaluated in an unsupervised way. For this purpose, we have proposed a continual setting, where hour-long videos are split into many 5s-long video clips that the model observes in sequence, one by one. All these short clips belong the the same video, which is continuous (has no cuts), and we call them chunks. At each step, whenever the network sees a new chunk, we randomly sample a chunk from anywhere in the whole video and pass it to the network as a query. This clip might belong to the part of the video that the network has already seen, i.e. its {\em past}, or a part that has not been yet observed, i.e. its {\em future}. We name our task {\em Past-Future}, and illustrate the whole setup in Figure~\ref{fig:online_learning_tasks}. As can be seen, to solve this task, the network can use memory of the past events. To avoid pure memorization, we always randomly and spatially crop the clips, so that the query is almost surely different than the past chunks. For training, we only do backpropagation over these chunks individually, and aggregate all such gradients together for the update step. In the following, we describe all the components, {\em memory, adapter, core} and {\em predictor}, that we use in our experiments on this task.
\paragraph{Memory.} 
Our task requires efficient memory usage. Here, we investigate that angle by comparing \textit{LSTM}, \textit{Slot} and using no memory (\textit{none}). A network without memory needs to respond to visual stimuli reactively, based only on the current observations. In our task, such network should perform at random chance (50-50), which is confirmed experimentally ($52.9\%$). \textit{LSTM} is the most popular recurrent neural network, equipped with gating operations that enable long short-term memory usage~\cite{hochreiter1997long}. 
Although LSTMs work well with shorter video sequences~\cite{donahue2015long,kim2018multi,li2018videolstm,srivastava2015unsupervised,sun2017lattice}, working with longer videos become more problematic due to  interference~\cite{aljundi2018memory,french1999catastrophic,goodfellow2013empirical,mccloskey1989catastrophic}. This is also the case here. The network is only able to learn the following simple strategy. On the question if the query is in the \textit{past}, it answers negatively for the first half of the video, and positively for the second half of the video. 
Note that, after observing almost the whole video the answer is very likely positive. This simple strategy yields $78.2\%$.
\textit{Slot} is an external memory unit~\cite{graves2014neural,oh2019video,weston2014memory,wu2019long} that can explicitly store past representations. In our study, we employ a deterministic writing operation, where a new memory is added to old memories whenever chunk $t$ is observed, i.e., $\mathcal{M}^{t+1} := \left\{\bs{m}^t\right\} \cup \mathcal{M}^t$.  $\mathcal{M}^t$ represents all the slots at time step $t$. The reading operation is learned using a specialized neural network. The network equipped with the \textit{slot} memory can solve the task on the half-an-hour long video understanding ($99.5\%$).
\paragraph{Adapter.} Adapter is a shallow 3D CNN that operates on a chunk creating the task-specific representation. It also reduces the dimensionality of the input. The same as other our experiments, chunks are already neural codes. Without compressing the inputs into such representations, it became challenging to even sample data points and transfer them through a bandwidth-limited network.
\paragraph{Core.} It is a neural network that interacts between queries and memories. Here, we experiment with a cross-source transformer, which uses the query as a transformer-query $q$ and memory elements as transformer-key $k$ and -values $v$ -- adopting naming convention from the transformer literature~\cite{vaswani2017attention} -- or it can also be seen as an asymmetric variant of a cross-modal transformer~\cite{tsai2019multimodal}. Note that, although standard transformer operate on the whole input sequence, here, the memory is a bottleneck between the input signal and the transformer. Moreover, due to the asymmetry, the network's cross-attention scales better than pure attention to longer sequences.
\paragraph{Predictor.} It is a single linear layer that outputs a scalar describing whether the query is in the \textit{past} or \textit{future}. Overall, to train networks on very long videos, we need to trade-off the network's complexity with its capacity. Thus, we use relatively simpler architecture like a shallow 3D CNNs in Adapter.

\section{Modelling}
\label{app:model}
This section discusses architectures and training protocols in more detail. In particular, we describe the recognition architecture, architectures used to implement \compressor and \augmentor, and our hyper-parameters.

\subsection{Architecture}

\subsubsection{S3D.}
\label{app:s3dapplied}
\begin{figure}
    \centering
    \includegraphics[width=\linewidth,page=1,trim={0cm 7cm 3.3cm 0cm},clip]{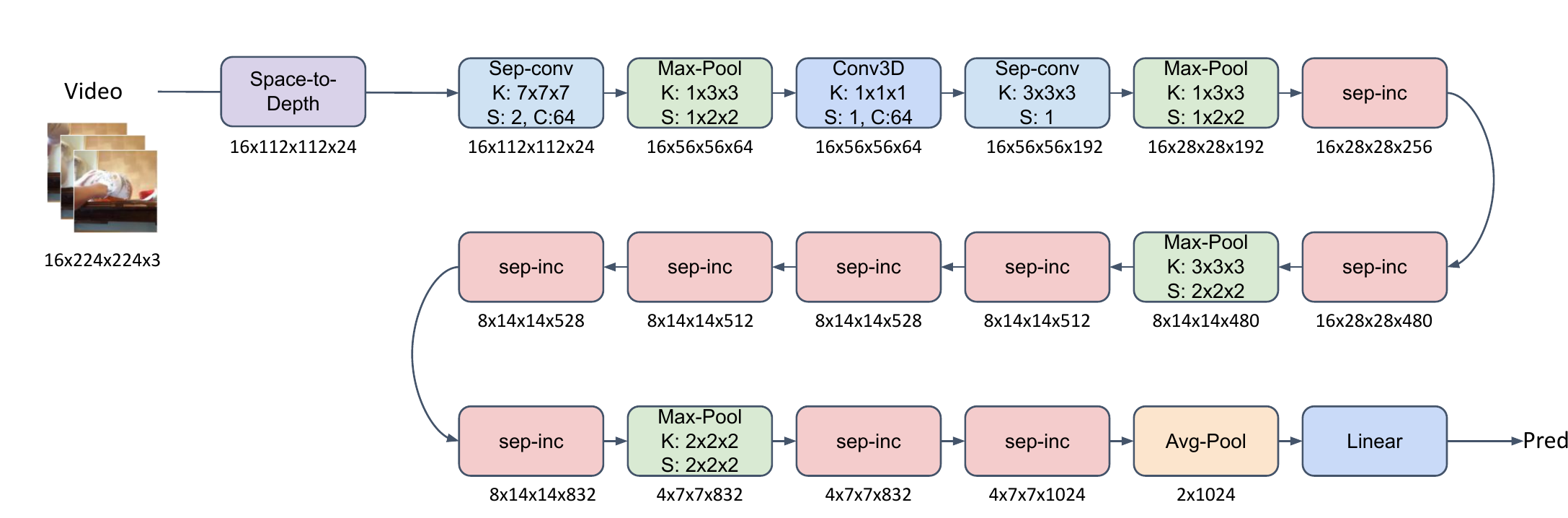}
    \caption{{\bf S3D.} Reproduction of Figure 6 from \cite{xie2017rethinking}. This shows how the standard S3D architecture is applied to a video. Note that we show the result after first applying a space to depth transformation to the input. Below each layer, we write the size of the output tensor for the given input size.}
    \label{app:fig:s3d}
\end{figure}

\begin{figure}
    \centering
    \includegraphics[width=\linewidth,page=2,trim={0cm 7cm 5.5cm 0cm},clip]{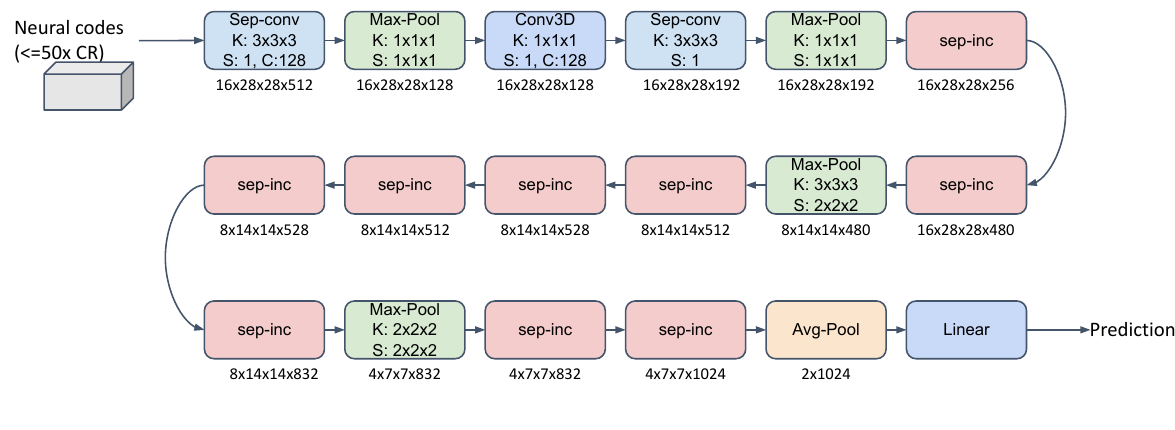}
    \caption{{\bf How we modify the standard S3D architecture for smaller compression rates.}. Below each layer, we write the size of the output tensor for the given input size. In comparison to Figure~\ref{app:fig:s3d}, we only change the strides of the first convolution, the first two max pools and modify the output channels in the first two convolutional layers.}
    \label{app:fig:smallcr}
\end{figure}

\begin{figure}
    \centering
    \includegraphics[width=\linewidth,page=3,trim={0cm 7cm 5.5cm 0cm},clip]{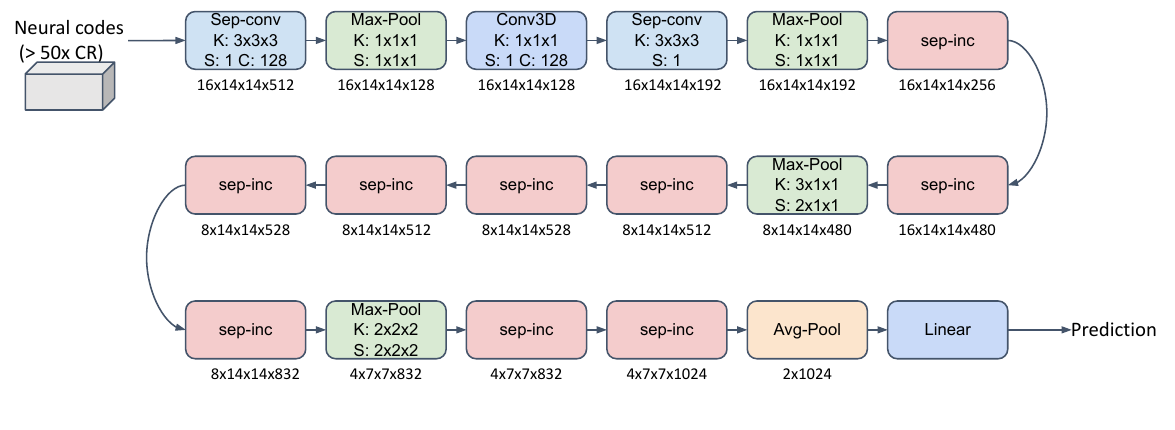}
    \caption{{\bf How we modify the standard S3D architecture for larger compression rates.}. Below each layer, we write the size of the output tensor for the given input size. In comparison to Figure~\ref{app:fig:s3d}, we only change the strides of the first convolution, the first three max pools and modify the output channels in the first two convolutional layers.}
    \label{app:fig:largercr}
\end{figure}

As compressed embeddings have smaller spatial dimensions than the RGB-valued inputs, we adapted the striding values of the S3D architecture so that the shapes of the internal tensors, between models operating on RGB and neural codes, are roughly the same. Figure~\ref{app:fig:s3d} reproduces Figure 6 from \cite{xie2017rethinking}; it shows how the S3D architecture is applied to a standard video input.
We then show the modifications to the strides and output channels that we use in S3D when operating on neural codes at CR$\sim30$ in the main paper; we visualise these changes in Figure~\ref{app:fig:smallcr}.
Figure~\ref{app:fig:largercr} visualises the changes if we have a larger compression rate and obtain codes that have a width and height of about one eighth the size of the original image. We use this setup for CR$\sim256$ and CR$\sim475$ in the main paper.
In general, applying S3D to our neural codes requires only a few small changes: modifying the strides of the input convolution and some of the max pool layers as well as modifying the output channels of the first two convolutional layers.

\clearpage

\subsubsection{Encoder-Decoder.}

The actual setup of the two components of the architecture is given in Table~\ref{tab:architecture}.
Note that in order to generalise to different numbers of frames, we have to reflect the tensor at the boundary when padding (as opposed to padding with zeros).

\paragraph{Encoder $c$.}
The encoder consists of $E_s$ ResNet blocks that make the resolution smaller.
This is followed by $E_c$ ResNet blocks that maintain the spatial resolution size.

\paragraph{Decoder $c^{-1}$.}
The decoder consists of $D_c$ ResNet blocks that maintain the spatial resolution size.
This is followed by $D_s$ ResNet blocks that increase the resolution to the original image resolution size.

\paragraph{Quantized embeddings.}
To create quantized embeddings, we take the $V_e$ channels and split them into $N$ codebooks. These are the embeddings we use to perform nearest neighbours and find the corresponding embedding in embedding space.

\begin{table}[]
    \centering
    \begin{tabular}{ccc}
        {\bf ID} & {\bf Output Size} & {\bf Block} \\ \toprule
        Encoder & $F\times H_c\times W_c \times V_e$ & $\begin{Bmatrix}
         \text{\underline{Conv3D}} \\ s=(1,2,2) \\ k=(4,4,4) \\oc=V_e / 2^{E_s - i - 1} \\ \text{ReLU} \\
         \end{Bmatrix} x E_s$ \\ \midrule
        Encoder & $F\times H_c\times W_c \times V_e$ & $\begin{Bmatrix}
         \text{\underline{Conv3D}} \\ s=(1,1,1) \\ k=(3,3,3) \\ oc=4V_e \\ \text{ReLU} \\
          \text{\underline{Conv3D}} \\ s=(1,1,1) \\ k=(3,3,3) \\ oc=V_e \\ \text{ReLU}
         \end{Bmatrix} x E_c$ \\ \midrule
        Decoder & $F\times H_c\times W_c \times V_e$ & $\begin{Bmatrix}
         \text{\underline{Conv3D}} \\ s=(1,1,1) \\ k=(3,3,3) \\ oc=4V_e \\ \text{ReLU} \\
          \text{\underline{Conv3D}} \\ s=(1,1,1) \\ k=(3,3,3) \\ oc=V_e \\ \text{ReLU}
         \end{Bmatrix} x D_c$ \\ \midrule
        Decoder & $F\times H \times W \times C$ & $\begin{Bmatrix}
         \text{\underline{Conv3DTranspose}} \\ s=(1,2,2) \\ k=(4,4,4) \\oc=V_e / 2^{i} \\ \text{ReLU} \\
         \end{Bmatrix} x (D_s - 1)$ \\ \midrule
        Decoder & $F\times H \times W \times C$ & $\begin{Bmatrix}
         \text{\underline{Conv3DTranspose}} \\ s=(1,2,2) \\ k=(4,4,4) \\oc=C \\
         \end{Bmatrix}$ \\ \midrule
         & 
    \end{tabular}
    \caption{{\bf Model Architecture.} The model architecture of the encoder-decoder models for an input video with $F$ frames of size $H, W, C$ ($C$ may be greater than $3$ if we are using spatio temporal crops). $s$ denotes stride while $k$ denotes the kernel size. $oc$ denotes the number of output channels and $i$ the block index within the group. }
    \label{tab:architecture}
\end{table}

\subsubsection{Augmentation network.}
\label{app:augmentor}
Our \augmentor has two main components: MLP and Transformer.
\paragraph{MLP.}
The MLP takes the input augmentation conditioning values, e.g. describing bounding box coordinates or brightness values or whether flipping happened, which are flattened to form a single vector; these are passed to three hidden layers of size $64$.
The output is a vector with the same number of channels as the neural code.
We broadcast the embedding from the MLP along the spatial dimensions and concatenate it with the neural code to give a tensor with double the size of the latter.

\paragraph{Transformer.}
The transformer takes as input the concatenated tensor.
The transformer has two hidden layers of size $128$, $4$ heads and uses an absolute positional embedding.

\subsection{VQ-VAE and Compression Rates}
\label{app:vqvaeandcr}
With the VQ-VAE encoding-decoding scheme, we can indirectly control  compression rates as follows. First, we use 3D CNNs with $L_c$ layers that downscale spatial dimensions of the input tensor $H_i \times W_i$ to form a tensor of shape $H_c\times W_c$. As we are using striding two, new dimension $H_c$ is $2^{L_c}$ times smaller than the corresponding dimension of the input frame. We do not compress along the time dimension. As we are using $N_c$ codebooks, the neural codes form a tensor with the shape $H_c\times W_c \times N_c$. Each element of such a tensor is a discrete number indicating which embedding in the corresponding codebook is used for the reconstruction. The larger codebook, the more bits per such an element are needed. Thus the final compression ratio is: $c_r = \frac{H_i\times W_i \times 3 \times \log_2(256)}{H_c \times H_w \times N_c \times \log_2(K_c)}$ where $K_c$ is the number of the codebook's elements.

\subsection{Sweeps}
\label{app:sweeps}

In order to choose the best hyperparameters for the compression architecture we swept over the following model choices. 
For a given setup, we chose the set of hyperparameters with the best performance on the validation set.

\subsubsection{Encoder-decoder architecture}

\paragraph{Input transformation.}
First, we swept over the type of spatio temporal patches.
We considered two setups: we can either use the original shape of the tensor or we can reshape the tensor by dividing it into temporal and spatial crops.
We considered either using the original video shape or dividing the video into four spatial crops and concatenating along the channel dimension.
However, we found that taking  further crops in either the spatial or temporal dimension hurt performance substantially, so did not investigate further.

\paragraph{Architecture size.}
We swept over the number of encoder ResNet blocks $E_s$ (we considered [3, 4, 5]) and corresponding decoder ResNet blocks $D_s$ (again [3, 4, 5]) that change the final spatial size of the embedding.
We also swept over the number of intermediary ResNet blocks $E_c, D_c$ in the encoder and decoder (we swept over using [3, 5, 7] blocks for both).
Here, we found as in \cite{mentzer2020high}, that using more decoder blocks improves reconstruction performance (as opposed to more encoder blocks).

\paragraph{Codebook.}
We swept over the number of codebooks (we considered one or two), the number of embeddings in the codebooks (we consider [256, 512, 1024, 4096, 8192]) and the size of those embeddings (we consider embeddings of size [128, 256, 512]).

\paragraph{Training parameters.}
Finally, we swept over the learning rate when training the encoder-decoder model. We considered learning rates [3e-4, 1e-5] and used an SGD optimizer.

\subsubsection{Augmentation Network}
\paragraph{MLP.}
We swept over the number of hidden dimensions and the size of those dimensions: we considered [[128], [128, 128], [64, 64, 64]]. In general, we found using more hidden layers performed better.

\paragraph{Transformer.}
We swept over the number of heads ([1, 4]) and the size of the intermediate representation ([128, 256]). In general, these choices did not make a large difference in the results.

\subsection{Training.}
\label{app:training}
Our pipeline consists of three training stages, which we detail below.
\paragraph{\bfcompressor.} The \compressor was trained until convergence on either the Kinetics600 or WalkingTours dataset.
To chose the best configuration of parameters, we swept over the hyperparameters (as described above) and selected the one with the best reconstruction loss on the validation set.
The precise sweeps are given in Section~\ref{app:sweeps}.
\paragraph{\bfaugmentor.} The  \augmentor was trained until convergence using a learning rate of $0.001$, the Adam optimizer and no weight decay. 
\paragraph{ {\bf Downstream training.}} Finally, the downstream networks were trained using the Adam optimizer with cosine decay and an initial learning rate of $0.5$ and weight decay of $1$e$-5$.
The models were trained for 60 epochs on COIN and 135 epochs on Kinetics600.

\bibliographystyle{splncs04}
\bibliography{egbib}

\end{document}